\documentclass[journal,12pt,onecolumn]{IEEEtran}  % 
\usepackage[margin=1in]{geometry}
\usepackage{setspace}
\usepackage[numbers]{natbib}
\usepackage{booktabs}
\usepackage{hyperref}
\usepackage{comment}
\usepackage{balance} 
\usepackage{amsmath,amssymb,amsthm}
\usepackage{color}
\usepackage{xcolor}
\usepackage{enumitem}
\usepackage{amsmath,amsfonts,bm}

\def\eqref#1{equation~\ref{#1}}
\def\1{\bm{1}}

\DeclareMathAlphabet{\mathsfit}{\encodingdefault}{\sfdefault}{m}{sl}
\SetMathAlphabet{\mathsfit}{bold}{\encodingdefault}{\sfdefault}{bx}{n}

\usepackage{xspace}
\usepackage{bbm}
\usepackage{mathrsfs}
\newcommand{\Ac}{\mathcal{A}}

\newcommand{\Ec}{\mathcal{E}}

\newcommand{\Hc}{\mathcal{H}}

\newcommand{\Rc}{\mathcal{R}}

\newcommand{\Uc}{\mathcal{U}}

\newcommand{\Rt}{{\widetilde{R}}}

\newcommand{\Vt}{{\widetilde{V}}}

\newcommand{\pt}{{\tilde{p}}}

\newcommand{\zt}{{\tilde{z}}}

\def\a{\alpha}

\def\l{\lambda}

\def\P{\mathrm{P}}

\def\textiid{i.i.d.\@\xspace}
\newcommand\iid{\ifmmode\text{ i.i.d. } \else \textiid \fi}

\newtheorem{theorem}{Theorem}
\newtheorem{corollary}{Corollary}

\newtheorem{lemma}{Lemma}
\newtheorem{proposition}{Proposition}
\newtheorem{remark}{Remark}
\theoremstyle{definition}
\newtheorem{assumption}{Assumption}

\makeatletter
\def\thickhline{%
  \noalign{\ifnum0=`}\fi\hrule \@height \thickarrayrulewidth \futurelet
   \reserved@a\@xthickhline}
\def\@xthickhline{\ifx\reserved@a\thickhline
               \vskip\doublerulesep
               \vskip-\thickarrayrulewidth
             \fi
      \ifnum0=`{\fi}}
\makeatother
\newlength{\thickarrayrulewidth}
\title{On the Adversarial Robustness of Learning-based Conformal Novelty Detection}

 \author{Daofu Zhang, Mehrdad Pournaderi, Hanne M. Clifford, 
\\Yu Xiang, Pramod K. Varshney\\
 \thanks{D. Zhang is with the Department of Electrical and Computer Engineering, University of Utah, email: daofu.zhang@utah.edu. M. Pournaderi is with Mofid Securities, email: M.Pournaderi@emofid.com. Y. Xiang is with the Department of Mathmatics and Statistics, Florida Atlanic University, email: yxiang@fau.edu. H. M. Clifford and P. K. Varshney are with the Department of Electrical Engineering and Computer Science, Syracuse University, email: \{hmsaarin, varshney\}@syr.edu. D. Zhang and Y. Xiang were supported in part by the National Science Foundation
under Grant CCF-2611415.}
}

\begin{document}

\maketitle

\begin{abstract}
  This paper studies the adversarial robustness of conformal novelty detection. In particular, we focus on two powerful learning-based frameworks that come with finite-sample false discovery rate (FDR) control: one is AdaDetect (by Marandon et al., 2024) that is based on the positive-unlabeled classifier, and the other is a one-class classifier-based approach (by Bates et al., 2023).  While they provide rigorous statistical guarantees under benign conditions, their behavior under adversarial perturbations remains underexplored. We first formulate an oracle attack setup, under the AdaDetect formulation, that quantifies the worst-case degradation of FDR, deriving an upper bound that characterizes the statistical cost of attacks. This idealized formulation directly motivates a practical and effective attack scheme that only requires query access to the output labels of both frameworks. Coupling these formulations with two popular and complementary black-box adversarial algorithms, we systematically evaluate the vulnerability of both frameworks on synthetic and real-world datasets. Our results show that adversarial perturbations can significantly increase the FDR while maintaining high detection power, exposing fundamental limitations of current error-controlled novelty detection methods and motivating the development of more robust alternatives.
\end{abstract}

\section{Introduction}
Consider the problem setup where training samples of \emph{typical} events are collected and used for detecting abnormal events from a large number of testing samples, where the \emph{abnormal} events follow a different distribution from the typical ones. This problem, known as \emph{novelty detection} (e.g.,~\cite{blanchard2010semi}), has attracted much attention recently through the lens of conformal $p$-values~\cite{vovk2005algorithmic,shafer2008tutorial,bates2023testing,mary2022semi,liang2022integrative,marandon2024adaptive,bashari2023derandomized}. The underlying metric is the false discovery rate (FDR)~\cite{benjamini1995,benjamini2001control,efron2001empirical,genovese2002operating,storey2002direct} that quantifies the false positives in a large-scale test dataset. Several frameworks have been developed with provable FDR control, requiring only exchangeability of the data under the null hypothesis. In particular, the ingenious method called \emph{AdaDetect}~\cite{marandon2024adaptive} by Marandon et al. introduces an adaptive transformation of detection scores, learned from both null and alternative samples, that yields finite-sample FDR guarantees under exchangeability. AdaDetect can be viewed as an important extension of several existing approaches, including the one-class classifier-based method by Bates et al.~\cite{bates2023testing} and also Bag of Null Statistics (BONus)~\cite{yang2021bonus}  (see Section~1.2 from~\cite{marandon2024adaptive} for detailed comparisons). 

The appealing properties of AdaDetect~\cite{marandon2024adaptive} and the one by Bates et al.~\cite{bates2023testing}, including the strong theoretical guarantees and efficient algorithm design, make them as potential strategies to empower existing safety-critical systems where training data samples are highly secure. For instance, in the banking system, customers' past personal transaction histories are highly protected yet can be used to enable fraud detection of new suspicious transactions when customers' account information is leaked and leveraged by malicious attackers. Thus, it is of great importance to quantify and evaluate the robustness of both approaches under various adversarial settings. In this work, we study the robustness of these two learning-based methods through the lens of adversarial machine learning that concerns the vulnerability of modern classifiers and detection systems under carefully designed perturbations. Importantly, we are interested in adversarial attacks \emph{directly on the data}, rather than on transformed scores (e.g., $p$-values) --- our study is the first of its kind in the literature, explicitly analyzing adversarial attacks on novelty detection systems while quantifying the cost in FDR control. Our proposed approach is flexible to incorporate existing adversarial machine learning attack algorithms. We hope to address the following two natural questions. 

\emph{How does a malicious attacker design an adversarial attack under FDR control?} As a first step in this direction, we focus on the setup where the test data can be attacked while keeping the training data intact, capturing the characteristics of safety-critical systems mentioned above. Specifically, we propose to first study the worst-case setting as a baseline to quantify the loss in FDR under the strongest possible attack. This formulation and the corresponding analysis provide critical guidelines for the design of practical attack schemes. Interestingly, we have developed a heuristic yet powerful algorithm that almost achieves the worst possible attack in our synthetic and real data experiments.

\emph{How to incorporate existing adversarial machine learning algorithms?} We propose a general methodology to make existing attack algorithms more effective. As a proof of concept, we adopt two popular off-the-shelf black-box adversarial attack algorithms, HopSkipJump~\cite{chen2020hopskipjumpattack} and Boundary Attack~\cite{brendel2018decision}, which only require the predicted label for refined data perturbation (see a detailed discussion in Section~\ref{sec:oracle}). These algorithms allow us to effectively change the score for the decision-making while minimizing the required changes in data. This approach enables directly attacks on the raw data, which is much more realistic than attacking scores of the data after some fixed transformation (e.g., $p$-values). For instance, there are works that focus on attacking $p$-values directly in the widely used Benjamini-Hochberg (BH) procedure for FDR control: a distributed setting using the BH procedure in the presence of compromised Byzantines~\cite{zhang2025distributed}, and on the adversarial robustness of the BH procedure through perturbation of $p$-values~\cite{chen2024adversarial}. For the seminal work by Chen et al.~\cite{chen2024adversarial}, see Appendix~\ref{app:compare} for a detailed comparison between their INCREASE-c method and our approach in  to demonstrate the perturbations required for effective attacks. 

\subsection{Contributions and Outline}

The main contributions of this work are threefold. First, we formulate and design adversarial attacks of the AdaDetect framework by proposing an oracle setting, when the attacker has access to both the model and test data labels, to quantify the upper bound on the loss in FDR (Section~\ref{sec:oracle}). Second, our oracle setting naturally motivates the design of a practical query-based attack scheme (Section~\ref{sec:surrogate}), called the surrogate decision-based attack, where the attacker can only query for the labels of the test data. This works for both AdaDetect and the one by Bates et al.~\cite{bates2023testing}. Third, in Section~\ref{sec:exp}, the vulnerability of AdaDetect and Bates et al.~\cite{bates2023testing} under our proposed attack strategies is extensively evaluated using two popular and complementary adversarial machine learning attack algorithms: HopSkipJump and Boundary Attack. The code can be found in the supplementary material.

%Here is the link to our simulation code: 
%\href{https://drive.google.com/file/d/1WZZTBU0qrDFLqPrdFaGYtT1GFK64QKbq/view?usp=drive_link}{Google drive repository}.

\subsection{Related Works}
\label{sec:related}

\noindent{\bf Novelty and anomaly detection with error control.}  
A growing body of work investigates conformal inference for novelty and anomaly detection with rigorous statistical guarantees~\cite{laxhammar2015inductive,smith2015conformal,ishimtsev2017conformal,guan2022prediction,cai2020real,haroush2021statistical}. While classical detectors~\cite{khan2014one,agrawal2015survey,chalapathy2019deep} often lack mechanisms for quantifying uncertainty, recent approaches provide explicit false discovery rate (FDR) control~\cite{yang2021bonus,bates2023testing,marandon2024adaptive}. Among these, AdaDetect~\cite{marandon2024adaptive} employs conformal $p$-values to guarantee FDR control while simultaneously learning the alternative distribution. Building on this line, \cite{bashari2023derandomized} introduce a conformal $e$-value framework that derandomizes novelty detection and achieves rigorous FDR guarantees. AutoMS~\cite{zhang2022automs} addresses model selection for out-of-distribution detection under controlled false discoveries, whereas online FDR-controlled anomaly detection~\cite{rebjock2021online} extends these guarantees to time-series data.  
These developments bring principled error control to novelty detection, but generally assume benign environments. Adversarial robustness in this context remains underexplored: \cite{lo2022adversarially} shows that one-class detectors are vulnerable to adversarial manipulation, yet without statistical error guarantees. This gap highlights the need to integrate FDR-controlled detection with robustness analysis against adversarial threats.

%\smallskip
%\subsection{Adversarial Attacks}
\noindent {\bf Adversarial machine learning.} Research on adversarial machine learning has revealed diverse classes of attacks depending on the adversary’s knowledge and resources.
In the {\bf white-box} setting, adversaries have full knowledge of the model, including parameters and gradients. Early gradient-based attacks include FGSM~\cite{goodfellow2015explaining}, BIM/I-FGSM~\cite{kurakin2017adversarial}, PGD~\cite{madry2018towards},
%, which iteratively refine perturbations. 
DeepFool~\cite{moosavi2016deepfool} and the CW attack~\cite{carlini2017towards}.
%formulate perturbations as optimization problems for minimal distortion.
JSMA~\cite{papernot2016limitations} reduces perturbations to only a few critical dimensions. Universal and generative attacks extend beyond instance-specific perturbations~\cite{moosavi2017universal,baluja2017adversarial}.
%: Universal Adversarial Perturbations (UAPs)~\cite{moosavi2017universal} create reusable noise vectors, while ATNs~\cite{baluja2017adversarial} train neural networks to generate adversarial examples.
More recent work considers spatial and semantic transformations, including Robust Physical Perturbations ($RP_2$)~\cite{eykholt2018robust} and spatially transformed adversarial examples~\cite{xiao2018spatially}. 
%These demonstrate that white-box adversaries can compromise models not only digitally but also in physical-world deployments. 
In the {\bf black-box} setting, adversaries lack direct access to model gradients or parameters. Instead, they rely on querying the model or leveraging transferability.
Score-based attacks estimate gradients using output probabilities, e.g., Zeroth-Order Optimization (ZOO)~\cite{chen2017zoo}, Natural Evolution Strategies (NES)~\cite{ilyas2018black}, and One-Pixel Attack~\cite{su2019one}. Decision-based attacks, such as the Boundary Attack~\cite{brendel2018decision}, HopSkipJump~\cite{chen2020hopskipjumpattack}, and Sign-OPT~\cite{cheng2020signopt}, require only the final predicted label and progressively refine perturbations.
Transfer-based methods exploit the phenomenon that adversarial examples often transfer across models: perturbations crafted on a surrogate can fool the target~\cite{papernot2016limitations}. %More recently, query-efficient black-box attacks (e.g., RayS; \cite{chen2020rays}) have demonstrated high success rates with few model queries, raising concerns about the practicality of adversarial robustness. 
See~\cite{zheng2025blackboxbench} for a comprehensive benchmark of black-box adversarial attacks. In response to the growing body of adversarial attacks, researchers have developed a range of defense mechanisms~\cite{xu2018feature,madry2018towards,zhang2019trades,cohen2019certified,papernot2016distillation,wong2018provable,metzen2017detecting}. Also see surveys \cite{wu2023attacks} and \cite{guo2025beyond} for such settings in computer vision.

\section{Background}

We have $n$ null training samples $\{Z_i\}_{i=1}^n$ sharing a common yet \emph{unknown} marginal distribution $P_0$ (also known as a semi-supervised setting~\cite{mary2022semi}), and $m$ unlabeled testing samples $\{Z_i\}_{i=n+1}^{m+n}$ where $m_0$ of the testing samples share the same distribution as $P_0$ while the rest $m_1=m-m_0$ of them follow different distributions. By convention, we will also refer to null samples as \emph{inliers} and non-null samples as \emph{outliers}. Let $\Hc_0$ contain all true null indices, while $\Hc_1$ contains all non-null indices in the testing data.
We define the key performance metrics as follows. Let $V$ be the number of true nulls that are incorrectly rejected (false discoveries) and $R$ the total number of rejections. 
The FDR is defined as
\begin{equation}
\text{FDR} = \mathbb{E}\left[\frac{V}{R \vee 1}\right],
\end{equation}
where $R \vee 1 := \max\{R, 1\}$. The \textbf{power} measures the detection performance of non-nulls, defined as $\text{power} = \mathbb{E}[(R - V)/(m_1 \vee 1)]$, where $m_1 = |\Hc_1|$ is the number of non-nulls in the test set. We write $[n_1:n_2]$, for $n_1< n_2$, to denote the set of consecutive numbers from $n_1$ to $n_2$, i.e., $\{n_1,n_1+1,..., n_2\}$.

\subsection{The AdaDetect Scheme~\cite{marandon2024adaptive}}\label{adascheme}
We use $\{X^\text{train}_1,\dots,X^\text{train}_n\}$ to represent null training samples and $\{X^\text{test}_1,\dots,X^\text{test}_m\}$ for the unlabeled testing samples. Following the notation from~\cite{marandon2024adaptive}, we combine them into $\{Z_i\}_{i=1}^{n+m}$ in this work where $\{Z_i\}_{i=1}^{n}$ represent $\{X^\text{train}_1,\dots,X^\text{train}_n\}$ and $\{Z_i\}_{i=n}^{n+m}$ represent $\{X^\text{test}_1,\dots,X^\text{test}_m\}$. Regarding the data generation mechanism, we make the following general assumption, which is the same as Assumption 1 of \cite{marandon2024adaptive}. 
   
% \begin{assumption}\label{ass1}
% Null data are exchangeable conditioned on the non-null data.
% \end{assumption}

\begin{assumption}[Exchangeability of nulls given non-nulls]\label{ass1}
\begin{equation*}
\begin{split}
&(Z_1, \ldots, Z_n, Z_{n+i} : i \in \mathcal{H}_0 ) \mid (Z_{n+j} : j \in \mathcal{H}_1) \overset{d}{=} \\
&\quad\quad\quad\quad(Z_{\pi(1)}, \ldots, Z_{\pi(n)}, Z_{\pi(n+i)} : i \in \mathcal{H}_0) \mid (Z_{n+j} : j \in \mathcal{H}_1)
\end{split}
\end{equation*}
for any permutation $\pi$ over $\{1, \ldots, n\} \cup \{n+i:i\in \Hc_0\}$.
\end{assumption}

AdaDetect is an adaptive novelty detection procedure that combines data-driven learning with distribution-free inference to provide finite-sample FDR control. It partitions the null sample into training data $\{Z_i\}_{i=1}^k$ and calibration data $\{Z_i\}_{i=k+1}^n$. The algorithm proceeds as follows.

\textbf{Step 1: Learn score function.} Partition the data into the training data $\{Z_i\}_{i=1}^k$ and the mixed sample $\{Z_i\}_{i=k+1}^{n+m}$, which contains both calibration samples from $P_0$ and unlabeled test samples. Under the \emph{positive-unlabeled (PU)} learning framework (labeling the first $k$ samples as null and the rest of them as non-nulls), apply a machine learning algorithm to learn a data-driven and measurable score function: $s: \mathcal{Z} \times \mathcal{Z}^k \times \mathcal{Z}^{n+m-k} \to \mathbb{R}$ as 
\begin{equation}\label{eq1}
s(z) := s(z; (Z_1, \ldots, Z_k), (Z_{k+1}, \ldots, Z_{n+m}))
\end{equation}
satisfying the following permutation property. For any permutation $\pi$ of $\{k+1, \ldots, n+m\}$,
\begin{align}
\label{eq:score}
&s(z; (z_1, \ldots, z_k), (z_{\pi(k+1)}, \ldots, z_{\pi(n+m)})) =  s(z; (z_1, \ldots, z_k), (z_{k+1}, \ldots, z_{n+m})).
\end{align}

\textbf{Step 2: Transform to scores.} Apply the learned function to obtain univariate scores such that a larger score indicates a higher likelihood of being a novelty, 
\begin{equation}
\begin{split}
&O_i = s(Z_i; (Z_1, \ldots, Z_k), (Z_{k+1}, \ldots, Z_{n+m})),  i \in [k+1: n+m].
\end{split}
\end{equation}

\textbf{Step 3: Compute conformal $p$-values.} For each test observation $Z_j$ with $j \in [n+1: n+m]$, generate conformal  $p$-values (also called empirical $p$-values) by comparing against the calibration set: for $j \in [1: m]$,
\begin{equation}
\begin{split}
&p_j = \frac{1}{n-k + 1} \left( 1 + \sum_{i=k+1}^n \mathbf{1}\{O_i > O_{n+j}\} \right).
\end{split}
\end{equation}

\textbf{Step 4: Apply BH procedure.} Apply the BH algorithm to $(p_{1}, \ldots, p_{m})$ to get the BH threshold $\tau$ at target level $\alpha$ and reject those $p$-values less than this threshold.

% We start with some notation. Let $\Ac$ be the set containing all attacked indices and $|\Ac|=m_{\Ac}$. 

\subsection{One-class classifier-based scheme by Bates et al.~\cite{bates2023testing}}\label{batescheme}
\label{sec:bates}
The authors in~\cite{bates2023testing} propose a  novel framework for outlier detection that can be implemented in two settings: a marginal setting, which is closely related to the AdaDetect scheme, and a calibration-conditional setting, which provides stronger guarantees for a fixed dataset. Both versions utilize a one-class classification model to produce $p$-values with finite-sample FDR control. This procedure with marginal $p$-values is directly analogous to AdaDetect, differing primarily in the score learning mechanism.

\textbf{Step 1: Learn score function.} A one-class classification algorithm (e.g., an Isolation Forest or One-Class SVM) is trained on $\mathcal{D}_{train}=\{Z_1,...,Z_k\}$ to learn a score function $s$. Unlike the PU learning in AdaDetect, this function is learned using only null samples.
    
\textbf{Step 2: Transform to scores.} The function $s$ is applied to the calibration set $\mathcal{D}_{cal}=\{Z_{k+1},...,Z_{n}\}$ and test samples to obtain univariate scores $O_i$, where smaller scores indicate a higher likelihood of being an outlier.
    
\textbf{Step 3: Compute marginal $p$-values.} For each test observation $Z_j$, an empirical $p$-value is generated based on its rank relative to the calibration set: for $j \in [1: m]$,
\begin{equation}
\begin{split}
    &p_j = \frac{1}{n-k + 1} \left(1 + \sum_{i=k+1}^n \mathbf{1}\{O_i > O_{n+j}\}\right).
    \end{split}
\end{equation}
    These $p$-values are marginally valid, meaning they control the error rate on average across different potential calibration sets.
    
\textbf{Step 4: Multiple testing.} For this marginal setting and a conditional setting mentioned below in Remark~\ref{remark:ccv}, the BH algorithm is applied to the resulting $p$-values at level $\alpha$. Because these $p$-values are proven to be positive regression dependent on a Subset (PRDS), the BH procedure maintains FDR control despite the shared calibration data.

\begin{remark}
\label{remark:ccv}
%\subsubsection{The Calibration-Conditional (MC) Setting}
To provide a guarantee for the specific calibration set $\mathcal{D}_{cal}$ held by the practitioner, Bates et al. introduce calibration-conditional validity (CCV). This ensures that with probability $1-\delta$, the $p$-values remain valid for the fixed data at hand. 

\textbf{Apply Monte Carlo adjustment after Step~3.} Marginal $p$-values $p_j$ are transformed into adjusted $p$-values $\hat{p}_j$ using a function $h$ determined through Monte Carlo simulation. 
    \begin{itemize}[nosep]
        \item The adjustment function $h$ is a simultaneous upper confidence bound for the empirical distribution, ensuring $p$-value validity at a confidence level of $1-\delta$.
        \item This MC approach is designed to preserve power by mimicking the Simes adjustment for small $p$-values while remaining efficient for larger values.
    \end{itemize}   
    \end{remark}
In Section~\ref{sec:exp}, we carry out experiments under both the marginal $p$-value and the CCV settings.

\section{Worst-case Attack and Practical Attack}
\label{sec:attacks}
In this section, we wish to study two attack schemes: an oracle attack and a query-based practical attack. Both schemes are compatible with existing black-box decision-based adversarial machine learning algorithms.

\subsection{Oracle Setting: Worst-case Attack Scheme}\label{sec:oracle}

% {\color{red}In this setting, assume the attacker has access to all null training samples $\{z_j\}_{j=1}^n$ and testing set, and he knows everything about the algorithm that the Adadetect is using. And he fits AdaDetect locally, obtaining the score function $s(z)$ . 
% }

 We first introduce an oracle setting to obtain an upper bound on the FDR loss when given the full information, enabling a theoretical analysis of the FDR behavior. We assume that the attacker has access to the full dataset with correct labels, as well as all the configurations of the algorithm used in AdaDetect by the user. Specifically, the attacker has
 
 {\bf Data:} Training samples $\{Z_j\}_{j=1}^n$ and test samples $\{Z_j\}_{j=n+1}^{m+n}$, and \emph{the attacker knows which test samples are nulls and non-nulls};
 
{\bf Algorithm:} All the information about AdaDetect implemented by the user, including the machine learning model for the score function and its parameters.

%AdaDetect's score function $s$. 
%and uses it for HopSkipJump.

%eliminating the need to train any surrogate classifier.

%\smallskip
We start by describing our first attack scheme (Step~1 and Step~2) and the outputs after applying AdaDetect directly on the attacked data (Step~3). 

\textbf{Step 1: Attack set selection.} Select a subset $\{Z_{n+i}: i\in \Ac\}$ from the true null test samples $\{Z_{n+i}: i\in \Hc_0\}$ as the attack target. We set the attack size as {\bf fixed size} where $|\Ac|=m_a$ for some fixed number $m_a$.

%, where the cardinality of the attack set is $|\Ac|=m_{\Ac}$. 

%We set the attack size as (1) {\bf fixed size} where $|\Ac|=m_a$ for some fixed number $m_a$, or (2) {\bf random size} where $m_{\Ac}$ is random. 

%$m_{\Ac} = \lfloor \gamma \cdot m_0\rfloor$, where $\gamma \in (0,1]$ is an ``attack intensity" parameter specified by the attacker.
%\begin{equation*}
%\ph_{(\tau_{\text{BH}}+1)}, \,,..., \,\ph_{(\tau_{\text{BH}}+m_{\Ac})}
%\end{equation*}

%\footnote{Our proofs in this work hold for any selection of such a set. See Remark~\ref{rem:select} for one natural and effective way to select the set, as demonstrated in our experiment section.}.

\textbf{Step 2: Decision-based adversarial perturbation.} Since the attacker has correct labels for all the data, the attacker can use them to train a score function $g(z)$ for the attack algorithm. 
  We form the labeled dataset
    \begin{equation*}
    \mathcal{D_\text{oracle}} = \{(Z_{i}, Y^*_i)\}_{i=1}^{m+n}
  \end{equation*} where $Y^*_i=0$ for $i\in\Hc_0$, $Y^*_i=1$ for $i\in\Hc_1$ are the \emph{true} labels.
  Then train a score function
  \begin{equation*}
    g(z) \leftarrow \mathrm{TrainScoreFunction}(\mathcal{D_\text{oracle}}).
  \end{equation*}For each $i\in \Ac$, generate
  \begin{align}
    \widetilde{Z}_{n+i} &=f_{\text{attack}}(Z_{n+i}\,;\,g(z))\nonumber\\   
    &:=f_{\text{attack}}( Z_{n+i}\,; \,\{Z_1,..., Z_n, Z_{n+j}: j\in  \Hc_0\setminus \Ac\},(Z_{n+j}:j\in A\cup \Hc_1))    \label{eq:attack}   
  \end{align} 
  such that $\mathbf{1}\{g(Z_{n+i})\ge0.5\}\ne \mathbf{1}\{g(\widetilde{Z}_{n+i})\ge0.5\}$, meaning that the decision is altered.
  We write 
  \begin{equation*}
      \{Z_1,..., Z_n, Z_{n+j}: j\in \Hc_0\setminus \Ac\}
  \end{equation*}      as an \emph{unordered} set to highlight that $f_{\text{attack}}$ does not depend on the order of elements in this set.  In our experiments (see Section~\ref{sec:exp}), we evaluate adversarial robustness using two representative decision-based attacks: the HopSkipJumpAttack (HSJA)~\citep{chen2020hopskipjumpattack} and the Boundary Attack~\citep{brendel2018decision}. Therefore, our attack scheme inherits their underlying optimized perturbation. In both HSJA and Boundary, they minimize $\lVert z- \zt\rVert^2$ such that the label is flipped, namely, $\mathbf{1}\{g(z)\ge0.5\}\ne \mathbf{1}\{g(\tilde{z})\ge0.5\}$.

\textbf{Step 3: Apply AdaDetect on the contaminated (or attacked) data.} After the attack, the user applies AdaDetect and computes the score function as the first step. As the data is now changed by the attacker, we denote the score function after the attack by $\tilde{s}(z)$, and the empirical $p$-values after the attack by $\pt_i$ for $i\in[1:m]$. We stress that $\tilde{s}(z)$ still satisfies~\eqref{eq:score}. The number of rejections is denoted by $\widetilde{R}(m_a)$, where we explicitly shows its dependence on $m_a$ for our main results in the next section.

The key in this oracle setting is that the attacker knows which ones are true nulls in the test data. The attacker will simply pick $\Ac$ with $m_{a}=|\Ac|$, which consists of a \emph{fixed} set of indices of nulls in the test data (i.e., there is no randomness in $\Ac$ and $m_a$). Our proposed methodology is flexible in that it can incorporate existing adversarial machine learning attack algorithms. 

%Consequently, this randomness propagates through all subsequent analysis: the cardinality $m_{\Ac} = |\Ac|$ is a random variable, and the intersection $\Ac \cap \Hc_0$ with the true null set is random. All of our results that involve $\Ac$ account for this randomness by considering the joint distribution of the whole dataset that induces the distribution of $\Ac$.

% \begin{remark}
% \label{rem:select}

% \end{remark}

The following proposition is critical for our analysis. It holds since (I) $f_\text{attack}(\cdot\,;\, g(z))$ only relies on the score function $g(z)$, and (II) $g(z)$ is invariant to order of elements in $\{Z_{1},\dots,Z_{n+i}:i\in\Hc_0\setminus \Ac\}$ as they are all labeled as $0$.
\begin{proposition}
\label{prop:perm}
$f_\text{attack}(\cdot\,;\, g(z))$ does not depend on the order of elements in $\{Z_{1},..., Z_n, Z_{n+j}: j\in \Hc_0\setminus \Ac\}$.
\end{proposition}

Decision-based attacks, including HSJA and Boundary attack, which rely solely on query access to the decision function, capture adversarial capabilities more realistically than white-box or score-based methods. Moreover, evaluating FDR loss under such attacks provides a stringent robustness assessment, as these ``blind" perturbations often induce more severe failures than those observed under other threat models. The two algorithms were chosen for their complementary properties. The Boundary Attack is a seminal decision-based approach that operates via a random-walk strategy, starting from an adversarial example and progressively reducing the perturbation while remaining misclassified. It is conceptually simple, model-agnostic, and widely adopted as a baseline in the literature. In contrast, HSJA is a more recent attack that achieves state-of-the-art query efficiency by combining binary search with adaptive estimation of the decision boundary's normal vector. While both attacks require only hard-label access to the model, Boundary Attack provides a robust baseline, whereas HSJA represents a stronger and more query-efficient adversary. Together, they allow us to assess robustness against both classical and modern decision-based adversarial paradigms.

 The attack in the last step pushes the score above the decision boundary. It is important to note that the output of the HSJA scheme~\cite{chen2020hopskipjumpattack} indeed does not depend on the order of elements in $\{Z_{1},..., Z_n, Z_{n+i}: i\in \Hc_0\setminus\Ac\}$. The same holds for the Boundary Attack~\citep{brendel2018decision}.

 \begin{remark}
     Although HSJA is sometimes described as a gradient estimation method, it does not require differentiability of the model; instead, it approximates the boundary’s normal vector using only hard-label queries. This makes it applicable even to non-differentiable classifiers such as random forests.
 \end{remark}

% \begin{remark}
%     We focus our analysis on attacks targeting the test data rather than the calibration data. Since the calibration data is sampled from the broader pool of available null observations, any adversarial attempt to compromise this set can be effectively neutralized by the server through a random re-partitioning of the training and calibration data.
% \end{remark}
 
% {\color{red} In this scheme, where does stability and power come into play? 
% \begin{itemize}
% \item We use $\tilde{s}(x)$ to denote the score function after the attack. Stability will make sure $\P\left(|s(x_i)-\tilde{s}(x_i)|>\epsilon\right)<\delta,i\notin \Ac$ and $\P\left(|s(z_j)-\tilde{s}(z_j)|>\epsilon\right)<\delta,1\leq j\leq n$. 
% \item Good detection power will make sure for each $i \in \Ac$, its underlying label is most likely to be true null.
% \end{itemize}
% }

% {\color{blue} Key questions:
% \begin{itemize}
%     \item In Thm 3.4, how would the definition of $p_i'$ change under this attack model? 
%     \item accordingly, how would the new $\hat{k}'_i$ related to $\hat{k}$? 
% \end{itemize}
% }
\subsection{Analysis}

In our oracle setting, we denote the corresponding $\text{FDR}$ as  $\text{FDR}^*_{\text{attack}}$. Our main theorem quantifies the loss in FDR caused by the attack. Recall that $\widetilde{R}(m_a)$ denotes the number of rejections when the user applies Adadetect on the contaminated data samples. The proof is deferred to Appendix~\ref{app:thm1}.
\begin{theorem}
\label{thm:1}
    Consider that $\Ac$ is a fixed set of indices with $m_{a}=|\Ac|$. Under Assumption~\ref{ass1}, with the score function $\tilde{s}$ satisfying the permutation invariance property in~\eqref{eq:score} and the attack scheme $f_{attack}$ being order-invariant as in~\eqref{eq:attack}, the FDR after the attack is
    \begin{equation}
\text{FDR}^*_{\text{attack}} \le \frac{m_0-m_a}{m}\alpha +m_{a}\cdot\mathbb{E}\left[\frac{1}{\Rt(m_a)\lor 1} \right],
\label{eq:thm1}
    \end{equation}where the expectations are taken over the randomness in the training and test samples $\{Z_j\}_{j=1}^{m+n}$.
    Furthermore, we have $\text{FDR}^*_{\text{attack}}\le \alpha$, as long as  
    \begin{equation}
    \label{eq:bound-ma}
    h(m_a)\le  \frac{m_1}{m}\,\a\,, \quad\text{ where }\quad
        h(x):= x \,\cdot \biggl(\mathbb{E}\left[\frac{1}{\Rt(x)\lor 1}\right]-\frac{\a}{m}\biggr).
    \end{equation}
\end{theorem}

It is straightforward to see that~\eqref{eq:bound-ma} comes from upper-bounding~\eqref{eq:thm1} by $\alpha$. As the bound on $m_a$ is implicit, we provide some clarifications to help gain a better understanding of the valid $m_a$'s that satisfies~\eqref{eq:bound-ma}. Consider a simple Gaussian setting (see details in Section~\ref{sec:exp}), for $m=1000$ with $m_1=100, 200, 300$, the corresponding ranges of $m_a$ are: $m_a=1$, $m_a\in [1:4]$, and $m_a\in [1:9]$, respectively.  As a rough approximation, one can even upper bound $\Rt(m_a)$ by $m$,  which gives $m_a\le \frac{\a}{1-\a}m_1$ (for instance, it gives $m_a\le 11$ for $m_1=100$ and $\alpha=0.1$), but it is important to keep in mind that \emph{this upper bound on $m_a$ no longer guarantees $\text{FDR}^*_{\text{attack}}\le \alpha$} and thus serves as a loose bound on the valid $m_a$ satisfying~\eqref{eq:bound-ma}. These all indicate that when $m_a$ is very small, AdaDetect is robust to any adversarial attacks. On the other hand, as detailed in our numerical experiments (Section~\ref{sec:exp}), we observe that $\text{FDR}^*_{\text{attack}}$ goes beyond $\alpha$ fairly quickly as $m_a$ grows. 

%It follows from Theorem~\ref{thm:1} and the simple fact that $\Rt_{m_a}\le m$.
% \begin{corollary}
%     Under the same setting as in Theorem~\ref{thm:1}, 
% \end{corollary}

Unlike the original AdaDetect, which guarantees $\text{FDR} \leq \alpha$ in benign settings, Theorem~\ref{thm:1} provides an upper-bound under adversarial perturbations. In our proofs, we start with decomposing the FDR into attacked and unattacked components, and our key technical innovations (Lemmas~\ref{lem:data} and~\ref{lem:score}) say that the first term $\mathbb{E}[\sum_{i \in \Hc_0 \setminus \Ac} \frac{\Vt_i}{{\Rt_{m_a}} \lor 1}]$ remains bounded by $\alpha$ even after perturbations, demonstrating that AdaDetect's control over unattacked samples is preserved.

\begin{remark}
    It turns out that the proof techniques for this oracle setting in Theorem~\ref{thm:1} can be adapted to less stringent settings where the true labels of test samples are unknown to the attacker (see Appendix~\ref{app:score}). %Specifically, using the training data $\{Z_j\}_{j=1}^n$ and the mixed sample $\{Z_j\}_{j=k+1}^{n+m}$, form the dataset $\mathcal{D} = \{(Z_i, Y_i)\}_{i=1}^{n+m}$, where $Y_i = 0$ for $i \in [1:k]$ and $Y_i = 1$ for $i \in [k+1:n+m]$ using the positive-unlabeled (PU) framework. 
\end{remark}

The key lemmas below show the conditional exchangeability we need for Theorem~\ref{thm:1}. First, we introduce the following notation for simplicity of presentation. Denote the number of \emph{unattacked true null test samples} by $\tilde{m}_0$. In order to simplify notation, let
\begin{align*}
   U_{\setminus\Ac} &= (U_1, \dots, U_{n + \tilde{m}_0}) :=(Z_1, \dots, Z_n, Z_{n+i} : i \in \Hc_0 \setminus \Ac), \\
    U_{\Ac}&=(U_{n + \tilde{m}_0+1}, \dots ,U_{n+m_0}):= (Z_{n+i} : i \in \Ac), \\ \widetilde{U}_{\Ac}&=(\widetilde{U}_{n + \tilde{m}_0+1}, \dots ,\widetilde{U}_{n+m_0}):= (\widetilde{Z}_{n+i} : i \in \Ac), \\
V &= (V_1, \dots, V_{m_1}) := (Z_{n+i} : i \in \Hc_1).
\end{align*}
With a slight abuse of notation, the condition~\eqref{eq:attack} on $f_\text{attack}(\cdot\,;\, g(z))$ can be simplified as 
  \begin{equation} 
    \widetilde{Z}_{n+i} = f_{\text{attack}}( Z_{n+i}\,;\, \,U_{\setminus\Ac},U_{\Ac}\cup V),
\end{equation}
where \emph{we stress that $f_{\text{attack}}$ does not depend on the order of the elements in $U_{\setminus\Ac}$}. Note that we have $Z_{n+i}=\widetilde{Z}_{n+i}$ for any $i\notin \Ac$. We will present two main lemmas that establish a crucial property that is unique to our adversarial setting: 
\begin{center}
    \emph{A form of conditional exchangeability is preserved even after adversarial perturbations.}
\end{center}
 %This is non-trivial because the attack set $\Ac$ is random and depends on all data through the BH procedure. 
 The key insight is that by conditioning on not only non-null data $V$ but also the attack outcomes $\widetilde{U}_{\Ac}$, the unattacked null samples maintain their exchangeable structure. 
 
  The main technical challenge arises from a certain form of ``matching" between the invariance property~(\eqref{eq:attack}) of $f_\text{attack}(\cdot\,;\, g(z))$ by the attacker and the invariance property~(\eqref{eq:score}) of score function $\tilde{s}(z)$ by the user on the contaminated data. (Recall from Step~3 of the oracle setting, $\tilde{s}$ satisfies~\eqref{eq:score} because of the PU classifier.) We explain this important point here before presenting the two lemmas. It turns out that for our analysis to work, we need to work with the exchangeability for the unattacked true null elements indexed from $k+1$ and onwards (i.e., $\{U_{k+1},\dots,U_{n+\tilde{m}_0}\}=\{Z_{k+1},..., Z_n, Z_{n+i}: i\in \Hc_0\setminus \Ac\}$) to derive an upper bound on FDR after attack. While the original AdaDetect analysis demonstrates exchangeability for all true null elements including the first $k$ training samples (i.e., $(Z_1, \ldots,Z_n, Z_{n+i} : i \in \Hc_0)$), this broader exchangeability does not hold in our adversarial setting due to the following subtle yet important consideration. 
  \begin{center}
   \emph{Both of the attack function $f_{\text{attack}}$  and the score function $\tilde{s}$  should be invariant w.r.t. $\{U_{k+1},\dots,U_{n+\tilde{m}_0}\}=\{Z_{k+1},..., Z_n, Z_{n+i}: i\in \Hc_0\setminus \Ac\}$.}
  \end{center}
   Fortunately, this holds because $g(z)$ is trained on true labels (assigning the same label to this set of data samples) and $f_\text{attack}(\cdot\,;\, g(z))$ depends on the data through $g(z)$, while $\tilde{s}(z)$ uses PU which also assigns the same label to this set, since PU labels the first $k$ samples $\{Z_i\}_{i=1}^k$ by $0$ and the rest of the samples $\{Z_i\}_{i=k+1}^{n+m}$ by $1$.  
   
   %See the proof in Appendix~\ref{app:lem1}. 

% {\color{red}The randomness of the attack set $\Ac$ does not affect this exchangeability proof because the conditional distribution statement holds for any fixed realization of $\Ac$. Specifically, once $\Ac$ is determined (which happens after the step $2$ in the attack scheme), the sets $U$, $U_{\Ac}$, $\widetilde{U}_{\Ac}$, and $V$ become well-defined, and the exchangeability property follows from the underlying distributional assumptions and the permutation-invariant structure of $f_{\text{attack}}$. }This preservation is essential for extending AdaDetect's theoretical framework to adversarial settings.

\begin{lemma}\label{lem:data}
    Under the setting of Theorem~\ref{thm:1}, we have 
    \begin{align*}
    \begin{split}
&(U_{k+1}, \dots, U_{n + \tilde{m}_0}) \mid V \cup \widetilde{U}_{\Ac} \notag \overset{d}{=}(U_{\pi(k+1)}, \dots, U_{\pi(n + \tilde{m}_0)}) \mid V \cup \widetilde{U}_{\Ac}       
    \end{split}
\end{align*}
for any permutation $\pi$ of the indices $\{k+1, \ldots, n + \tilde{m}_0\}$.
\end{lemma}

\begin{proof}

From Assumption~\ref{ass1}, for any permutation $\pi$ of  the indices $\{1, \ldots, n + \tilde{m}_0\}$ such that $\pi(i) = i$ for $i \leq k$, we have
\begin{equation}\label{lm1eq10}
(U_{\setminus\Ac} \mid V) \overset{d}{=} (U_{\setminus\Ac}^{\pi} \mid V).
\end{equation}
From the property of the attack algorithm, we have for each $i\in \Ac$ that
\begin{equation}
    \widetilde{Z}_{n+i}=f_{\text{attack}}(Z_{n+i} \,; \,U_{\setminus\Ac},U_{\Ac}\cup V), 
\end{equation}
and we write this in a compact form as $\widetilde{U}_{\Ac}:=f_{\text{attack}}(U_{\Ac}\,; \, U_{\setminus\Ac},U_{\Ac}\cup V)$. 
% \begin{equation}
%     \widetilde{U}_{\Ac}=f_{\text{attack}}(\{\widetilde{Z}_{n+i}: i\in \Ac\}\,; \,U,U_{\Ac},V)
% \end{equation}
% $\widetilde{U}_{\Ac}=f_{\text{attack}}(U,U_{\Ac},V)$, where $f_{\text{attack}}$ does not depend on the order of elements in $U$ as in~\eqref{eq:attack} and here we write $f_{\text{attack}}$.
We want to show that 
\begin{align}
    &(U_{\setminus\Ac}\mid f_{\text{attack}}(U_{\Ac}\,;\, U_{\setminus\Ac},U_{\Ac}\cup V),V)\overset{d}{=}(U_{\setminus\Ac}^{\pi}\mid f_{\text{attack}}(U_{\Ac}\,; \, U_{\setminus\Ac}^{\pi},U_{\Ac}\cup V),V).
\end{align}
 According to Proposition~\ref{prop:perm},  $f_{\text{attack}}$ does not depend on the order of the elements from $U_{k+1}$ to $U_{n+\tilde{m}_0}$ in $U_{\setminus\Ac}$, so we have $\widetilde{U}_{\Ac} = f_{\text{attack}}(U_{\Ac}\,; \,U_{\setminus\Ac}^{\pi},U_{\Ac}\cup V)$. For any measurable set $\Uc_{\setminus\Ac}$ in the support of $U_{\setminus\Ac}$, we have
\begin{align*}
&\P\left(U_{\setminus\Ac} \in \Uc_{\setminus\Ac} \mid \widetilde{U}_{\Ac} = \tilde{u}_{\Ac},V = v\right) =
\frac{\P(U_{\setminus\Ac} \in \Uc_{\setminus\Ac},  f_{\text{attack}}(U_{\Ac}; U_{\setminus\Ac},U_{\Ac}\cup V)=\tilde{u}_{\Ac},\ V=v)}{\P( f_{\text{attack}}(U_{\Ac}\,;\, U_{\setminus\Ac},U_{\Ac}\cup V)=\tilde{u}_{\Ac},\ V=v)}.
\end{align*}
According to (\ref{lm1eq10}), we know that all the elements inside $U_{\setminus\Ac}$ are exchangeable given $V$. Along with the assumption that $f_{\text{attack}}$ satisfies~\eqref{eq:attack}, we have that
\begin{align*}
&\P(U_{\setminus\Ac} \in \Uc_{\setminus\Ac}, f_{\text{attack}}(U_{\Ac}; U_{\setminus\Ac},U_{\Ac}\cup V)=\tilde{u}_{\Ac} | V=v)
\\
&\quad\quad\quad=\P(U_{\setminus\Ac}^{\pi} \in \Uc_{\setminus\Ac}, f_{\text{attack}}(U_{\Ac}; U_{\setminus\Ac}^\pi,U_{\Ac}\cup V)=\tilde{u}_{\Ac} | V=v).
\end{align*}
Therefore,
\begin{align}
&\P\left(U_{\setminus\Ac} \in \Uc_{\setminus\Ac} \mid \widetilde{U}_{\Ac} = \tilde{u}_{\Ac}, V = v\right) = \P\left(U_{\setminus\Ac}^{\pi} \in \Uc_{\setminus\Ac} \mid \widetilde{U}_{\Ac} = \tilde{u}_{\Ac}, V = v\right),
\end{align}
i.e., $\{U_{k+1},\dots,U_{n+\tilde{m}_0}\}$ is conditionally exchangeable given $(\widetilde{U}_{\Ac},V) = ( f_{\text{attack}}(U_{\Ac}\,;\, U_{\setminus\Ac},U_{\Ac}\cup V),V)$.
\end{proof}

    %Specifically, their exchangeability expression covers $(Z_1, \ldots,Z_n, Z_{n+i} : i \in \Hc_0)$ conditioned on $V$, while our restricted result only requires $(Z_{k+1}, \ldots,$
    %$ Z_n, Z_{n+i} : i \in \Hc_0 \setminus \Ac )$ to be exchangeable conditioned on $\widetilde{U}_{\Ac}$ and $V$.

Now we show that the conditional exchangeability of data in Lemma~\ref{lem:data} can be carried over to the scores, building on the key property that both of $f_{\text{attack}}$ and $\tilde{s}$ are invariant w.r.t. $\{U_{k+1},\dots,U_{n+\tilde{m}_0}\}=\{Z_{k+1},...,Z_n ,Z_{n+i}:i\in\Hc_0\setminus A\}$, as we alluded to before. %See the proof in Appendix~\ref{app:lem2}. 
\begin{lemma}\label{lem:score}
    Under the setting of Lemma~\ref{lem:data} and assume that $\tilde{s}$ satisfies~\eqref{eq1}, then we have \begin{align*}
&(\tilde{s}(U_{k+1}), \ldots,\tilde{s}(U_{n + \tilde{m}_0})) \mid (\tilde{s}(Z_{n+j}) : j \in \mathcal{H}_1)\cup(\tilde{s}(\widetilde{Z}_{n+j}) : j \in \mathcal{A})\\ 
&\quad\quad\quad\quad\overset{d}{=} (\tilde{s}(U_{\pi(k+1)}), \ldots, \tilde{s}(U_{\pi(n+\tilde{m}_0)})) \mid(\tilde{s}(Z_{n+j}) : j \in \mathcal{H}_1)\cup(\tilde{s}(\widetilde{Z}_{n+j}) : j \in \mathcal{A})
\end{align*}
for any permutation $\pi$ of the indices $\{k+1, \ldots, n + \tilde{m}_0\}$.
\end{lemma}

\begin{proof}
First, we introduce the following notion $Q$ to capture not only the invariance property in~\eqref{eq:score} satisfied by $\tilde{s}$, but also the invariance property in~\eqref{eq:attack} of $f_{\text{attack}}$, 
\begin{align}
    Q &:= h(U_{\setminus\Ac},\widetilde{U}_{\Ac},V)\nonumber\\
    &\overset{(a)}{=}((Z_1,\dots ,Z_k), \{Z_{k+1},...,Z_n ,Z_{n+i}:i\in\Hc_0\setminus A\}, (Z_{n+i}, i\in \Hc_1\cup \Ac)\,),
\end{align} 
where $(a)$ follows from (1) the key property that $f_{\text{attack}}$ and $\tilde{s}$ are invariant w.r.t. $\{Z_{k+1},...,Z_n ,Z_{n+i}:i\in\Hc_0\setminus A\}$, and (2) recall that $\widetilde{U}_{\Ac}:=f_{\text{attack}}(U_{\Ac}\,; \, U_{\setminus\Ac},U_{\Ac}\cup V)$. By Lemma~\ref{lem:data}, we have
\begin{equation}
(U_{\setminus\Ac},\widetilde{U}_{\Ac}, V) \overset{d}{=} (U_{\setminus\Ac}^\pi,\widetilde{U}_{\Ac}, V),
\end{equation}
with a slight abuse of notation where $\pi$ is defined over $[1: n + \tilde{m}_0]$ such that $\pi(i)=i$ for $i\le k$.

Since $Q$ is a function of $(U_{\setminus\Ac},\widetilde{U}_{\Ac},V)$, we have
\begin{align}
&(U_{\setminus\Ac},\widetilde{U}_{\Ac}, V, Q) = (U_{\setminus\Ac},\widetilde{U}_{\Ac}, V, h(U_{\setminus\Ac},\widetilde{U}_{\Ac}, V)) \overset{d}{=} 
(U_{\setminus\Ac}^\pi, \widetilde{U}_{\Ac},V, h(U_{\setminus\Ac}^\pi,\widetilde{U}_{\Ac}, V)).
\end{align}
Since $\pi$ keeps the first $k$ indices fixed, by the definition of $Q$, we have $h(U_{\setminus\Ac}^\pi,\widetilde{U}_{\Ac}, V) = h(U_{\setminus\Ac},\widetilde{U}_{\Ac}, V) = Q$. Thus
\begin{equation}
(U_{\setminus\Ac}, \widetilde{U}_{\Ac},V, Q) \overset{d}{=} (U_{\setminus\Ac}^\pi,\widetilde{U}_{\Ac}, V, Q).
\end{equation}

Applying the score function $\tilde{s}$ to each $U_i$ in $U_{\setminus\Ac}$, we obtain
\begin{align}
&(S_1, \dots, S_{n + \tilde{m}_0}) \mid \widetilde{U}_{\Ac},V, Q \overset{d}{=} 
(S_{\pi(1)}, \dots, S_{\pi(n + \tilde{m}_0)}) \mid \widetilde{U}_{\Ac},V, Q\,\,.
\label{eq:score-invariance}
\end{align}
Observe that the score terms $(\tilde{s}(Z_{n+j}) : j \in\Hc_1) \cup (\tilde{s}(\widetilde{Z}_{n+j}): j \in \Ac )$ are deterministic functions of the conditional variables $\widetilde{U}_{\Ac},V, Q$; recall that $Z_{n+i}=\widetilde{Z}_{n+i}$ for any $i\notin \Ac$.
Since $\pi(i) = i$ for all $i \leq k$, the exchangeability is preserved as follows, 
\begin{align*}
&(S_{k+1}, \dots, S_n, S_{n+i} : i \in \Hc_0 \setminus \Ac) \mid (\tilde{s}(Z_{n+j}) : j \in \Hc_1) \cup (\tilde{s}(\widetilde{Z}_{n+j}): j \in \Ac )  \\
&\quad\quad\overset{d}{=}(S_{\pi(k+1)}, \dots,S_{\pi(n)},S_{\pi(n + i)} : i \in \Hc_0 \setminus \Ac) \mid (\tilde{s}(Z_{n+j}) : j \in \Hc_1) \cup (\tilde{s}(\widetilde{Z}_{n+j}): j \in \Ac ).
\label{eq:null-score-exchange}
\end{align*}

Thus
$(\tilde{s}(Z_{k+1}), \dots, \tilde{s}(Z_n), \tilde{s}(Z_{n+i}) : i \in \Hc_0 \setminus \Ac)$ is exchangeable conditional on $(\tilde{s}(Z_{n+j}) : j \in \Hc_1) \cup (\tilde{s}(\widetilde{Z}_{n+j}): j \in \Ac )$.

\end{proof}

Building on Lemma~\ref{lem:data} and Lemma~\ref{lem:score}, the following result follows directly from~\cite[Theorem~A.1~(iii) and~(iv)]{marandon2024adaptive} and we skip the proof. 

%include the proof of it in Appendix~\ref{app:lem3} for completeness. 

\begin{lemma}
\label{lem:prop}
Under the Lemma~\ref{lem:score} setting. Let $i \in \Hc_0$ be an unattacked null index and  $S_i:=\tilde{s}(\widetilde{Z}_i)$ for all $i$. Define
% \begin{align*}
% \widetilde{W}_i 
%   \;=\;
%   &(\{S_{k+1},\dots,S_n,S_{n+i}\}
%   \;,\;(S_{n+j}: j\neq i, j\in \Hc_0)\;,(S_{n+j}:j\in \Hc_1)\,).     
% \end{align*} 
\begin{align*}
\widetilde{W}_i 
  \;=\;
  &(\{S_{k+1},\dots,S_n,S_{n+i}\}
  ,\;(S_{n+j}: j\neq i, j\in \Hc_0\setminus \Ac)\;,\;\\ &\hspace{12em}(S_{n+j}:j\in \Hc_1)\,\cup (S_{n+j}: j\neq i, j\in \Ac).     
\end{align*}

Then we have,

(i) The $p$-value $\pt_i$ is independent of $\widetilde{W}_i$.
  
(ii) The quantity $(n-k + 1)\pt_i$ is uniformly distributed on the integers $\{1, \dots, n-k + 1\}$.

\end{lemma}

\subsection{Surrogate Decision-based Attack Scheme}
\label{sec:surrogate}
Motivated by our oracle setting, we consider the following practical scenario. The attacker does not have the training samples $\{Z_j\}_{j=1}^n$ and does not know the true labels of the test samples, but is allowed to query the label from the user who applies AdaDetect, with underlying machine learning algorithms unknown to the attacker. Such query access is a standard assumption in adversarial settings as mentioned in Section~\ref{sec:related}, reflecting a realistic constraint on the attacker’s capability. Specifically, the attack has

{\bf Data:} Only test samples $\{Z_j\}_{j=n+1}^{m+n}$, but the attacker does not know which ones are nulls and non-nulls;

{\bf Query:} The attacker can query the user (who owns all the training and test data and the AdaDetect algorithm) to obtain the labels for all the testing data $\{Z_{n+j}\}_{j=1}^{m}$. We denote these labels as $\{Y_i\}_{i=1}^m$ and refer to them as pseudo-labels, since they are not necessarily the true label in contrast to $Y_i^*$ in the oracle setting.

We propose a \emph{surrogate score function} $g_{\text{surrogate}}(z)$, trained on the pseudo-labeled dataset
$\mathcal{D}_{\text{surroage}} = \{(Z_{n+i}, Y_i)\}_{i=1}^m$. (Unlike the true labels available in our oracle setting, these are the labels assigned by AdaDetect on the entire test set.) We now describe the three steps of this attack method. 

%This surrogate score function approximates AdaDetect's binary decision: 
% \[
% \mathbf{1}\{ g_{\text{surrogate}}(Z_{n+i})\ge0.5\}\approx Y_i^*,
% \]
% enabling decision-based adversarial attacks on a black-box detector.

\textbf{Step 1: Initial detection.} The attacker queries the labels of the test data from the user. Upon request, the user applies AdaDetect to the full test set $\{Z_{n+i}\}_{i=1}^m$ at once, producing pseudo-labels $Y_i\in\{0,1\}$ where, 
  \begin{equation*}
    (Y_1,\dots,Y_m) = \mathrm{AdaDetect}\left(\{Z_{n+i}\}_{i=1}^{m}\right).
 \end{equation*}

\textbf{Step 2: Surrogate score function training.} Assuming AdaDetect has reasonable detection power, we form the pseudo-labeled dataset $\mathcal{D}_{\text{surrogate}} = \{(Z_{n+i}, Y_i)\}_{i=1}^m$
  %   \begin{equation*}
  %   \mathcal{D} = \{(Z_{n+i}, Y_i)\}_{i=1}^m
  % \end{equation*}
and train a surrogate score function
  \begin{equation*}
    g_{\text{surrogate}}(z) \leftarrow \mathrm{TrainScoreFunction}(\mathcal{D}_{\text{surrogate}}).
  \end{equation*}

\textbf{Step 3: Attack set selection and adversarial perturbation.} Select a subset $\mathcal{A}$ from the unrejected test samples as the target, where $|\Ac|=m_a$ for some fixed number $m_a$.
  For each $i\in \mathcal{A}$, compute
  \begin{equation}
    \widetilde{Z}_{n+i} = f_\text{attack-surroage}(Z_{n+i}\,; \,g_{\text{surroage}}(z)).\label{eq:su_attack}
    \end{equation}

\begin{remark}
    One natural choice of $\Ac$ is to select the unrejected hypotheses with the smallest $p$-values (i.e., those closest to the rejection boundary). Let $(i_1, i_2, \ldots, i_{m-R})$ denote the indices of unrejected hypotheses ordered by their $p$-values: $\hat{p}_{i_1} \leq \hat{p}_{i_2} \leq \cdots \leq \hat{p}_{i_{m-R}}$. Then we define
$\Ac = \{i_1, i_2, \ldots, i_{m_{\Ac}}\}$. This selects the $m_{\Ac}$ unrejected indices with the smallest $p$-values, targeting hypotheses that are close to $\hat\tau$ and making it an effective attack strategy as demonstrated in our experiments.

\end{remark}

For this surrogate method, the main challenge in analyzing the FDR is because~\eqref{eq:attack} no longer holds, which implies that $f_{\text{surrogate-attack}}$ does not satisfy Proposition~\ref{prop:perm}. Specifically, recall that in the oracle setting,  $g(z)$ is invariant to order of elements in $\{Z_{1},\dots,Z_{n+i}:i\in\Hc_0\setminus \Ac\}$ as they are all labeled as $0$, which are the true labels known to the attacker. However, this fails to hold in the surrogate setting because (I) $g_{\text{surrogate}}(z)$ is trained only on the test data, and more importantly, (II) $g_{\text{surrogate}}(z)$ is trained on the pseudo-labels rather than the true labels, which breaks the invariance property.  

  % \begin{equation}
  %   \widetilde{Z}_{n+i} = f_\text{surrogate-attack}(Z_{n+i}, g_{\text{surroage}}(z)).\label{eq:su_attack}
  %   \end{equation}

% \begin{align}
%     \widetilde{Z}_{n+i} &=f_{\text{surrogate-attack}}(Z_{n+i};\,g(z))\nonumber\\   
%     &:=f_{\text{surrogate-attack}}( Z_{n+i}; \,\{Z_1,..., Z_n, Z_{n+j}: j\in  \Hc_0\setminus \Ac\},(Z_{n+j}:j\in A\cup \Hc_1))      
% \end{align}

It is important to note that our surrogate decision-based attack does not require information about the algorithm, making it a practical attack scheme. This point is also highlighted in our experiment section through mismatched setups, where the user and attacker adopt two different algorithms to learn the score function (see Experiment~3 and Experiment~A.1 for details).

\begin{remark}
We note that another possible attack is to treat AdaDetect as a black-box and apply a decision-based attack directly to its outputs. However, this might require a prohibitively large query budget, since changing the empirical FDR demands perturbing many test samples, and each sample in turn requires multiple queries to attack successfully.
\end{remark}

% \subsubsection{Analysis}
% The only difference from previous section is the score function (classifier). Previous score function is obtained from AdaDetect procedure which we conclude that the $f_\text{attack}$ does not depend on order of elements in $U$ (because of the PU labeling and null data are from the same distribution). Here we are using the surrogate classifier as the parameter for the $f_\text{attack}$, the question is whether $f_\text{attack}$ still does not depend on the order of elements in $U$. The surrogate classifier is trained using the predicted label along with test data from AdaDetect. So if we permute $U$, the training data and label for the surrogate classifier might change a lot. 

\section{Experiments}
\label{sec:exp}

As explained in Section~\ref{sec:oracle}, we will focus on two adversarial machine learning attacks: HSJA and Boundary attack. Recall that they minimize $\lVert z- \zt\rVert^2$ such that the label is flipped, which is inherited in all of our experimental evaluations. Each experiment is repeated $20$ times to calculate the mean and variance of the FDR and power. We denote the AdaDetect score function by $\tilde{s}(z)$ and the score function used for attack by ${g}(z)$ for oracle and $g_{\text{surrogate}}(z)$ for our surrogate method. The estimated upper bound according to~\eqref{eq:thm1} is computed by $\frac{m_0-m_a}{m}\alpha + m_a \cdot \frac{1}{20} \sum_{i=1}^{20} \left(\frac{1}{\Rt^{(i)} \lor 1}\right)$. This is computed for most of the experiments and listed in the tables.  
In Appendix~\ref{app:exp}, we present the synthetic data generation and experiment results, along with one more real-world data experiment.

\noindent{\bf Real-world datasets:} 
\begin{itemize}[nosep]
\item \textbf{Shuttle:} Radiator data onboard space shuttles. Instances from class $1$ are considered nominal, while instances from classes $2$--$7$ are considered novelties \cite{Dua2019}.
 
\item\textbf{Credit Card:} Transactions made by credit cards, some of which are fraudulent \cite{DalPozzolo2015}.

\item\textbf{KDDCup99:} Network connections that includes a variety of simulated intrusions \cite{KDD99}.

\item\textbf{Mammography:} Features from mammograms and some with microcalcifications \cite{Woods1993}.

\end{itemize}

\subsection{Applying Surrogate Scheme on Adadetect}
We evaluated the attack performance on three types of synthetic data distributions: independent Gaussian, non-Gaussian, and exchangeable Gaussian data, and four real-world datasets (Shuttle, Credit Card, KDDCup99, and Mammography) with diverse characteristics and application domains. All experiments use the following base parameters: training sample size $n=5000$, testing sample size $m=1000$, $k=4000$, true null data $m_0=900$, and significance level $\alpha=0.1$. Each experiment is repeated $20$ times to calculate the mean and variance of the FDR and power.

\smallskip
\noindent{\bf Experiment~1: Varying attack size with random forest (RF) models.}
We evaluate the impact of attack size on both oracle and surrogate attack performance using identical RF architectures for both score functions. The attack size $m_a$ sweeps between $10$, $30$, $50$ and $200$. We assess how each attack scheme performs under different attack scales with homogeneous RF configurations. 

We only test HSJA in synthetic data experiments while both of HSJA and Boundary attack are evaluated in real-world data experiments. The result shows that the FDR for both attack schemes are under the theoretical upper bound, and oracle attack outperforms surrogate attack in non-gaussian case when the original power is low. Overall, both attack schemes successfully increase the FDR across varying attack sizes.

{\renewcommand{\arraystretch}{1.5}
\begin{table*}[h!]
\centering
\caption{Experiment~1: FDR + RF}
\renewcommand{\arraystretch}{1.0}
\begin{tabular}{lcccc}
\hline
\hline
{\bf Dataset} & {\bf Independent Gaussian} & {\bf Non-Gaussian} & {\bf Exchangeable Gaussian} \\
\hline\hline
original FDR    &$0.08 \pm 0.03$  & $0.08 \pm 0.04$ &$0.08 \pm 0.04$   \\
  \hline
oracle ($m_a=10$)          &$0.11 \pm 0.02$  & $0.08 \pm 0.05$ & $0.10 \pm 0.01$\\
oracle ($m_a=30$)          &$0.24 \pm 0.02$  & $0.10 \pm 0.06$ & $0.28 \pm 0.01$\\
  \hline
oracle ($m_a=50$)          &$0.36 \pm 0.02$  & $0.40 \pm 0.05$ & $0.38 \pm 0.02$\\
surrogate ($m_a=50$) & $0.34 \pm 0.02$ &$0.20 \pm 0.05$  & $0.37 \pm 0.02$ \\
\hline
oracle ($m_a=200$)          & $0.67 \pm 0.00$ &$0.71 \pm 0.01$   & $0.69 \pm 0.00$    \\
surrogate ($m_a=200$)& $0.67 \pm 0.00$ &$0.64 \pm 0.01$   & $0.67 \pm 0.00$   \\
\hline\hline
\end{tabular}

\end{table*}
}

%\vspace{-2em}
{\renewcommand{\arraystretch}{1.5}
\begin{table*}[h!]
\centering

\caption{Experiment~1: Power + RF}
\renewcommand{\arraystretch}{1.0}
\begin{tabular}{lcccc}
\hline
\hline
{\bf Dataset} & {\bf Independent Gaussian} & {\bf Non-Gaussian} & {\bf Exchangeable Gaussian} \\
\hline\hline
original power   &$0.96 \pm 0.02$  & $0.55 \pm 0.06$ &$1.00\pm 0.00$   \\
  \hline
oracle ($m_a=50$)           &$0.99 \pm 0.01$  & $0.87 \pm 0.01$ & $1.00 \pm 0.00$\\
surrogate ($m_a=50$) & $0.96 \pm 0.02$ &$0.65 \pm 0.05$  & $1.00 \pm 0.00$ \\
oracle ($m_a=200$)          & $0.99 \pm 0.01$ &$0.98 \pm 0.01$   & $1.00\pm 0.00$    \\
surrogate ($m_a=200$)& $0.96 \pm 0.02$ &$0.78 \pm 0.05$   & $1.00 \pm 0.00$   \\
\hline\hline
\end{tabular}

\end{table*}
}

\smallskip
\noindent{\bf Experiment~2: Real-world data with RF models.}
We evaluated both oracle and surrogate attack performance on the four real-world datasets using identical RF architectures for both score functions, with attack size $m_a=200$. This allows us to compare attack effectiveness across different real-world data characteristics. We compare the boundary attack with our default HSJA under both oracle and surrogate attack schemes, using identical RF architectures for both score functions with $m_a=200$. The result shows that our oracle and surrogate attack schemes and decision-based algorithms (HSJA and Boundary) can significantly increase the FDR in comparison to the original FDR. We also report the simulation with $m_a=10$ for the surrogate method with Boundary attack in Table~IV.

{\renewcommand{\arraystretch}{1.5}
\begin{table*}[h!]
\centering
\caption{Experiment~2: FDR + RF ($m_a=200$)}
\renewcommand{\arraystretch}{1.0}
\begin{tabular}{lcccc}
\hline
\hline
{\bf Dataset} & {\bf Credit-card} & {\bf Shuttle} & {\bf KDD} & {\bf Mammography} \\
\hline\hline
original FDR   &$0.08 \pm 0.03$  & $0.01 \pm 0.00$ &$0.04 \pm 0.02$  & $0.04 \pm 0.08$  \\
  \hline
oracle+{\bf hop.}          &$0.60 \pm 0.02$  & $0.65 \pm 0.02$ & $0.48 \pm 0.10$ & $0.51 \pm 0.10$\\
surrogate+{\bf hop.} & $0.56 \pm 0.02$ &$0.66 \pm 0.03$  & $0.45 \pm 0.08$ &$0.45 \pm 0.11$ \\
oracle+{\bf bound.}          & $0.61 \pm 0.02$ &$0.68 \pm 0.02$   & $0.65 \pm 0.07$   & $0.61 \pm 0.05$ \\
surrogate+{\bf bound.}& $0.64 \pm 0.03$ &$0.70 \pm 0.02$   & $0.67 \pm 0.06$   & $0.57 \pm 0.04$ \\
\hline
estimated upper bound& $0.85$ &$0.73$   & $0.69$   & $0.88$ \\
\hline\hline
\end{tabular}

\end{table*}
}
{\renewcommand{\arraystretch}{1.5}
\begin{table}[h!]
\centering
\caption{Experiment~2: FDR + RF ($m_a=10$)}
\renewcommand{\arraystretch}{1.0}
\begin{tabular}{lcccc}
\hline
\hline
{\bf Dataset} & {\bf Credit-card} & {\bf Shuttle} & {\bf KDD} & {\bf Mammography} \\
\hline\hline
original FDR   &$0.08 \pm 0.03$  & $0.01 \pm 0.00$ &$0.04 \pm 0.02$  & $0.04 \pm 0.08$  \\
  \hline
surrogate+{\bf bound.}& $0.10 \pm 0.01$ &$0.09 \pm 0.01$   & $0.09 \pm 0.01$   & $0.07 \pm 0.05$ \\
\hline
estimated upper bound& $0.18$ &$0.18$   & $0.18$   & $0.20$ \\
\hline\hline
\end{tabular}

\end{table}
}
%\vspace{-2em}
{\renewcommand{\arraystretch}{1.5}
\begin{table*}[h!]
\centering

\caption{Experiment~2: Power + RF}
\renewcommand{\arraystretch}{1.0}
\begin{tabular}{lcccc}
\hline
\hline
{\bf Dataset} & {\bf Credit-card} & {\bf Shuttle} & {\bf KDD} & {\bf Mammography} \\
\hline\hline
original power    &$0.78 \pm 0.03$   & $0.84 \pm 0.02$ & $0.88 \pm 0.04$ &$0.48 \pm 0.09$ \\
  \hline
  oracle+{\bf hop.} &$0.86 \pm 0.03$  &$0.99 \pm 0.01$  &  $0.94 \pm 0.05$  & $0.67 \pm 0.07$\\
surrogate+{\bf hop.}          &$0.87 \pm 0.03$  &$0.99 \pm 0.01$  & $0.93 \pm 0.05$ &  $0.80 \pm 0.05$\\

oracle+{\bf bound.}          &$0.98 \pm 0.02$   & $0.97 \pm 0.02$ &$0.95 \pm 0.01$  &$0.80 \pm 0.09$ \\
surrogate+{\bf bound.} &$0.95 \pm 0.03$  &$0.96 \pm 0.03$  &$0.96 \pm 0.01$  & $0.78 \pm 0.10$\\
\hline\hline
\end{tabular}

\end{table*}
}

We make two important observations. Firstly, compared to the other three datasets, the original power on Mammography is relatively low ($\sim 0.48$). As a consequence, the surrogate method learns the score function from less accurate labels, leading to a larger gap in FDR between the oracle vs. the surrogate. Similar phenomena appear in the next real-world experiments, as well as the two in the Appendix~\ref{app:exp}. Secondly, the power after attack often increases, since the attack targets data points near the decision boundary in practice, some of which belong to the alternative hypothesis and are thus more likely to be correctly rejected.

\smallskip
\noindent{\bf Experiment~3: Real-world data with mismatched configurations.}
We apply mismatched score function configurations (RF--NN) to the four real-world datasets with a fixed attack size of $m_a = 200$, evaluating both the oracle and surrogate attack performance. This enables us to assess how different model combinations affect each attack type's performance across various real-world scenarios. The results indicate that an attacker can employ a model different from the user's and still inflate the FDR beyond the target level. Our experiments show that using the neural network configuration as the attacker’s model can substantially speed up the attack process. The results for RF-RF configuration is covered in Experiment $2$.
{\renewcommand{\arraystretch}{1.5}
\begin{table*}[h!]
\centering
\caption{Experiment~3: FDR + RF-NN}
\renewcommand{\arraystretch}{1.0}
\begin{tabular}{lcccc}
\hline
\hline
{\bf Dataset} & {\bf Credit-card} & {\bf Shuttle} & {\bf KDD} & {\bf Mammography} \\
\hline\hline
original FDR   &$0.09 \pm 0.05$  &$0.01 \pm 0.01$  & $0.02 \pm 0.01$ &$0.09 \pm 0.05$  \\
  \hline
oracle+{\bf bound.}          &$0.64 \pm 0.03$  & $0.69 \pm 0.02$ &  $0.69 \pm 0.02$&$0.69 \pm 0.01$ \\
surrogate+{\bf bound.}  &$0.60 \pm 0.02$  & $0.50 \pm 0.03$ &  $0.67 \pm 0.03$&$0.64 \pm 0.01$ \\
\hline
estimated upper bound& $0.79$ &$0.72$   & $0.69$   & $0.85$ \\
\hline\hline
\end{tabular}

\end{table*}
\begin{table*}[h!]
\centering
\caption{Experiment~4 with Monte Carlo adjustment}
\renewcommand{\arraystretch}{1.0}
\begin{tabular}{lccc}
\hline
\hline
{\bf Dataset} & {\bf Credit-card} & {\bf Shuttle} & {\bf Gaussian} \\
\hline\hline
original FDR          & $0.09 \pm 0.02$ & $0.08 \pm 0.04$ & $0.10 \pm 0.02$ \\
\hline
surrogate+{\bf hop.}   & $0.45 \pm 0.08$            & $0.58 \pm 0.08$             & $0.56 \pm 0.09$              \\
surrogate+{\bf bound.}& $0.41 \pm 0.10$ & $0.53 \pm 0.07$ & $0.69 \pm 0.06$ \\
\hline\hline
\end{tabular}
\end{table*}
}

\begin{table*}[h!]
\centering
\caption{Experiment~4 without Monte Carlo adjustment}
\begin{tabular}{lccc}
\hline
\hline
{\bf Dataset} & {\bf Credit-card} & {\bf Shuttle} & {\bf Gaussian} \\
\hline\hline
original FDR          & $0.08 \pm 0.01$ & $0.09 \pm 0.02$ & $0.09 \pm 0.03$ \\
\hline
surrogate+{\bf hop.}   & $0.45 \pm 0.08$            & $0.49 \pm 0.10$             & $0.58 \pm 0.07$              \\
surrogate+{\bf bound.}& $0.48 \pm 0.07$ & $0.54 \pm 0.06$ & $0.65 \pm 0.09$ \\
\hline\hline
\end{tabular}
\end{table*}

%\vspace{-2em}

% \paragraph{Experiment~7: Different score attack implementations.}
% We compare boundary attack with our default HopSkipJump attack under both oracle and surrogate attack schemes, using identical RF architectures for both score functions with $m_a=200$. Ideally, both boundary attack and HopSkipJump attack should be evaluated in each of the settings above; we report results for this representative setup due to computational constraints. 
% \bigskip
% {\color{red} In this scheme, where does stability and power come into play? 
% \begin{itemize}
%     \item ?
% \end{itemize}
% }
\subsection{Adapting Our Surrogate Approach to Bates el al.~\cite{bates2023testing} as in Section~\ref{sec:bates}}
To demonstrate the generality of our attack framework, we evaluate the adversarial robustness of the one-class classifier-based method proposed by Bates et al. in~\cite{bates2023testing}. This approach serves as a fundamental baseline. We utilize two real-world datasets and a synthetic Gaussian data distribution. The null (e.g., non-fraud) data is partitioned into a training set of size $n_{\text{train}}=4000$ and a calibration set of size $n_{\text{cal}}=1000$. The test set $\mathcal{D}_{\text{test}}$ consists of $m=2000$ samples, comprising $m_0=1800$ inliers (normal transactions) and $m_1=200$ outliers (e.g., fraudulent transactions). The $p$-values are computed using one-class classifier-based scheme, and the BH procedure is applied for FDR control.

\noindent\textbf{Surrogate Decision-based Attack Scheme.} We implement the surrogate decision-based attack scheme described in Section~\ref{sec:surrogate} with specific adaptations:

 \textbf{(1) Target Selection:} The attacker selects a set $\mathcal{A}$ of size $m_a=200$ from the unrejected test samples. To maximize attack efficiency, we select $200$ unrejected test samples with the smallest $p$-values, corresponding to samples naturally close to the decision boundary.

 \textbf{(2) Surrogate Training with Label Flipping:} The attacker queries the one-class classifier-based scheme to obtain initial binary rejections. To train the surrogate model $g_{\text{surrogate}}(z)$ (a Multi-Layer Perceptron with one hidden layer of 100 units), we employ a heuristic label-flipping strategy: $75\%$ of the rejected samples with the highest $p$-values are relabeled as inliers in the training set. This encourages the surrogate to learn a more restrictive decision boundary, as only the most extreme outliers retain their original labels. Since HSJA and Boundary attack are decision-based attack, this surrogate model with tighter decision boundary for outliers will encourage those two decision-based attack to be more aggressive. We remark that a more conservative boundary (e.g., using a $50\%$ threshold or no flip at all) often fails here because it results in a loose rejection region, making it difficult for the attack to successfully push a sample labeled inlier into the real outlier space. The sample may successfully move into this loose boundary for outliers, but it's still not extreme enough for one-class classifier-based scheme to notice. While the optimal threshold may vary across different datasets, this parameter serves as a tuning knob: increasing the flipping percentage makes the HSJA and Boundary attack more aggressive, facilitating more effective adversarial generation.

\textbf{(3) Adversarial Generation:} We utilize HSJA and Boundary attack on the surrogate model $g_{\text{surrogate}}(z)$ to generate adversarial perturbations for the target set $\mathcal{A}$. The attack is aiming to cross the decision boundary. 

\smallskip
\noindent{\bf Experiment~4: Both real-world and synthetic data on Bates's method.}
We evaluated the attack performance at significance levels $\alpha = 0.1$ both with and without Monte Carlo adjustment (see Remark~\ref{remark:ccv}). The results quantify whether our surrogate approach succeed at attacking the standard one-class classifier-based scheme. 

\section{Discussion}

We believe that this work opens up a wide range of possible directions concerning the interplay between adversarial robustness and conformal novelty detection. We briefly comment on two potential research directions. 

{\bf Defense and robust training.} In response to the growing body of research on adversarial attacks, researchers have developed a range of defense mechanisms. Early approaches focused on input preprocessing, such as feature squeezing~\cite{xu2018feature} or randomized transformations, but these were often circumvented by adaptive adversaries. More principled methods emphasize robust training. Adversarial training~\cite{madry2018towards} has become the de facto standard, where models are trained on adversarial examples generated during training to improve robustness. %Variants include TRADES~\cite{zhang2019trades}, which balances accuracy and robustness via a regularized objective, and randomized smoothing~\cite{cohen2019certified}, which provides certified robustness guarantees against $L_2$ perturbations. 
In the context of novelty detection with FDR control, these techniques suggest potential defenses against adversarially induced FDR inflation: robust training can make the decision boundary less susceptible to small perturbations, while randomized smoothing could stabilize conformal scores or $p$-values, thereby preserving statistical error guarantees under attack. Exploring such defenses offers a promising direction for integrating adversarial robustness with principled error control. 

% Other strategies include defensive distillation~\cite{papernot2016distillation}, certified defenses using convex relaxations\cite{wong2018provable}, and detection-based defenses that aim to reject adversarial inputs rather than classify them~\cite{metzen2017detecting}. Despite significant progress, a key challenge remains: many defenses provide robustness in restricted threat models (e.g., bounded $L_p$ perturbations), leaving models vulnerable to more diverse, semantic, or physical-world attacks.

{\bf Attack on training or calibration data.} As a first step in understanding the robustness of AdaDetect, we consider the security-critical scenarios where the training data is highly secure. It would be interesting to study the impact of attacks on the null samples, including the training and calibration data. This can be a suitable setup for less powerful agents, such as power-limited sensors or local servers in decentralized formulations (e.g.~\cite{zhang2025distributed,pournaderi2023sample}). For instance, consider that each sensor is deployed in the environment for monitoring, then attacking the calibration data is more reasonable and powerful, as it changes the reference for all the test samples. 

%  \section{Acknowledgment}
% Daofu Zhang and Yu Xiang were supported in part by the National Science Foundation
% under Grant CCF-2611415.

\balance
\bibliography{ref}
\bibliographystyle{IEEEtran}

\newpage
\onecolumn
\appendices

\section{Proof of Theorem~\ref{thm:1}}\label{app:thm1}

With Lemma~\ref{lem:data} and Lemma~\ref{lem:score} in place, the proof of Theorem~\ref{thm:1} largely follows from that of~\cite[Theorem 4.3]{marandon2024adaptive}. We present it here for completeness.

\begin{proof}[Proof of Theorem \ref{thm:1}]
Let $\Vt_i$ denote the indicator function for the rejection of hypothesis~$i$ and $\tilde{\tau}$ be the BH threshold under the adversarial attack, where 
\begin{equation}
\Vt_i = \mathbf{1}\{\,\pt_i \le \alpha\,(\tilde{\tau}/m)\} .
\end{equation}
We can decompose $\text{FDR}^*_{\text{attack}}$ as follows,
    \begin{equation*}
\hspace{-0.1em}\text{FDR}^*_{\text{attack}}\hspace{-0.2em}=\hspace{-0.2em}\mathbb{E} \left[ \sum_{i \in \Hc_0 \setminus \Ac} \frac{\Vt_i}{{\Rt_{m_a}} \lor 1} \right] + \mathbb{E} \left[ \sum_{i \in \Ac\cap \Hc_0} \frac{{\Vt}_i}{\Rt_{m_a} \lor 1} \right].
    \end{equation*}
    
Let $\Rt_{m_a} = \sum_{i=1}^{m} \Vt_i$ be the total number of rejections after the attack.
For the second term, we can bound it by
\begin{equation}
\mathbb{E} \left[ \sum_{i \in \Ac\cap \Hc_0} \frac{{\Vt}_i}{\Rt_{m_a} \lor 1} \right] \leq \mathbb{E} \left[\frac{|\Ac\cap \Hc_0|}{\Rt_{m_a}\lor 1} \right]\overset{(a)}{=} \mathbb{E} \left[\frac{m_a}{\Rt_{m_a}\lor 1} \right],\label{eq:up}
\end{equation}
where $(a)$ follows since $\Ac\subseteq \Hc_0$. For the first term, define $S_i=\tilde{s}(\widetilde{Z}_i)$ for $i\in [1:m+n]$. Fix any $i \in \Hc_0 \setminus \Ac$, and for $j \neq i$, we have
\begin{equation*}
  C_{i,j}
  =
  \frac{1}{n-k+1}
  \left(
   \sum_{s \,\in\,\{S_{k+1},\dots,S_n,\,S_{n+i}\}} 
      \mathbf{1}\{\,s > S_{n+j}\}
  \right).
\end{equation*}
Define the empirical $p$-values after attack
  \begin{align*}  
     \pt_i = \frac{1 + \sum_{j=k+1}^n \mathbf{1}\{\tilde{s}(Z_j) > \tilde{s}(\widetilde{Z}_{n+i})\}}{n-k + 1}
     =\frac{1 + \sum_{s \,\in\,\{S_{k+1},\dots,S_n\}} \mathbf{1}\{s > S_{n+i}\}}{n-k + 1}.
\end{align*}
We now create the \emph{auxiliary} p‐value vector $(p'_1,\dots,p'_m)$ by
\begin{equation}
  p'_j = 
  \begin{cases}
    \dfrac{1}{n-k+1} & \text{if } j = i, \\
    C_{i,j} & \text{if } j \neq i.
  \end{cases}
\end{equation}
By construction, we have $p'_j \le \pt_j$ whenever $\pt_j \le \pt_i$, since replacing $S_{n+j}$ with a smaller score $S_{n+i}$ yields a smaller count. On the other hand, if $\pt_j > \pt_i$, it follows that $p'_j = \pt_j$. Also, when $i=j$, $p_j'=1/(n-k+1)$ is the smallest possible value. This means that Condition~(63) in Lemma~D.6 from~\cite{marandon2024adaptive} is satisfied for \emph{all} $(i,j)$.
Recall $\tilde{\tau}$ is the BH index for $(\pt_1,\dots,\pt_m)$ and let $\tau_i':=\tau'_{\text{BH}}$ be the BH index for $(p'_1,\dots,p'_m)$. By Lemma~D.6 from ~\cite{marandon2024adaptive}, we obtain
\begin{equation}
  \left\{\pt_i \le \alpha\,(\tfrac{\tilde{\tau}}{m})\right\}
  =
  \left\{\pt_i \le \alpha\,(\tfrac{\tau'_i}{m})\right\}
  \;\subseteq\;
  \{\tilde{\tau} = \tau'_i\}.
\end{equation}
Focusing on $i \in \Hc_0 \setminus \Ac$, we have 
\begin{equation}
  \mathbf{1}\left\{\pt_i \le \alpha\,(\tilde{\tau}/m)\right\}
  =
  \mathbf{1}\left\{\pt_i \le \alpha\,(\tau'_i/m)\right\}.
\end{equation}
Summing over $i\in \Hc_0 \setminus \Ac$,
\begin{align*}
\mathbb{E} \left[ \sum_{i \in \Hc_0 \setminus \Ac} \frac{\Vt_i}{{\Rt_{m_a}} \lor 1} \right]
  &=
  \mathbb{E}\left[\sum_{i \in \Hc_0 \setminus \Ac}
  \frac{\,\mathbf{1}\{\,\pt_i \le \alpha\,(\tilde{\tau}/m)\}}{\tilde{\tau}}\right]
  =\mathbb{E}\left[\sum_{i \in \Hc_0 \setminus \Ac}
  \frac{\,\mathbf{1}\{\,\pt_i \le \alpha\,(\tau'_i/m)\}}{\tau'_i}\right].
\end{align*}
We define 
% \begin{align*}
%   &\widetilde{W}_i 
%   \;=\;(
%   \{S_{k+1},\dots,S_n,S_{n+i}\}
%   \;,\;
% (S_{n+j}: j\neq i, j\in H_0)
%   \;,\;
%   (S_{n+j}:j\in H_1)\,),
% \end{align*}
\begin{align*}
\widetilde{W}_i 
  \;=\;
  &(\{S_{k+1},\dots,S_n,S_{n+i}\}
  ,\;(S_{n+j}: j\neq i, j\in \Hc_0\setminus \Ac)\;,\;\\ &\hspace{12em}(S_{n+j}:j\in \Hc_1)\,\cup (S_{n+j}: j\neq i, j\in \Ac)\,).     
\end{align*} 
and note that $\tau'_i$ is \emph{$\widetilde{W}_i$-measurable}.
Hence
\begin{align*}
\mathbb{E} \left[ \sum_{i \in \Hc_0 \setminus \Ac} \frac{\Vt_i}{{\Rt_{m_a}} \lor 1} \right]
  &=\mathbb{E}\left[\sum_{i \in \Hc_0 \setminus \Ac}
  \mathbb{E}\left[
    \frac{\mathbf{1}\left\{\pt_i \le \alpha\,(\tau'_i/m)\right\}}{\tau'_i}
    \,\Big|\,
    \widetilde{W}_i
  \right]\right]\\
  &=
  \mathbb{E}\left[\sum_{i \in \Hc_0 \setminus \Ac}
  \frac{1}{\tau'_i}\,
  \mathbb{E}\left[
    \mathbf{1}\left\{\pt_i \le \alpha\,(\tau'_i/m)\right\}
    \,\Big|\,
    \widetilde{W}_i
  \right]\right],    
\end{align*}
where the last equality is due to $\tau'_i$ acting as a known constant conditional on $\widetilde{W}_i$.
From Lemma~\ref{lem:prop}, we know that $(n-k+1)\pt_i$ is the rank of $S_{n+i}$ among 
$\{S_{k+1},\dots,S_n,S_{n+i}\}$ and $\pt_i$ is independent of $\widetilde{W}_i$.  
As a result, $(n-k+1)\,\pt_i$ is uniform 
over $[1:n-k+1]$, independent of $\tau'_i$. Thus
\begin{align*}
  &\mathbb{E}\left[
    \mathbf{1}\{\,\pt_i \le \alpha\,(\tau'_i/m)\}
    \,\Big|\,
    \widetilde{W}_i
  \right]
  =\mathbb{P}\left(
     (n-k+1)\,\pt_i \;\le\; \alpha\,(n-k+1)\,\tfrac{\tau'_i}{m}
     \,\Big|\,
     \widetilde{W}_i
  \right)
  =\frac{
    \lfloor \alpha(n-k+1)\,\tfrac{\tau'_i}{m}\rfloor
  }{n-k+1}.
\end{align*}
Hence
\begin{align}
&\mathbb{E} \left[ \sum_{i \in \Hc_0 \setminus \Ac} \frac{\Vt_i}{{\Rt_{m_a}} \lor 1} \right]=
  \mathbb{E}\left[\sum_{i \in \Hc_0 \setminus \Ac}
    \frac{
       \lfloor \alpha(n-k+1)\,\tau'_i / m\rfloor
    }{
       (n-k+1)\,\tau'_i
    } \right]\le \frac{m_0-m_a}{m}\alpha.
\end{align}
This completes the proof since~\eqref{eq:bound-ma} comes from upper-bounding~\eqref{eq:thm1} by $\alpha$.

\end{proof}

\section{Direct Decision-based Attack Scheme}\label{app:score}

% {\color{red}In this setting, assume the attacker has access to all null training samples $\{z_j\}_{j=1}^n$ and testing set, and he knows everything about the algorithm that the Adadetect is using. And he fits AdaDetect locally, obtaining the score function $s(z)$ . 
% }
Different from the oracle setting, we assume that the attacker has access to

 {\bf Data:} Training samples $\{Z_j\}_{j=1}^n$ and test samples $\{Z_j\}_{j=n+1}^{m+n}$, but \emph{the attacker does not know which test samples are nulls and non-nulls};
 
{\bf Algorithm:} All the information about the AdaDetect implemented by the user, including the machine learning model for the score function and its parameters.

With such information at hand, the attacker is able to apply AdaDetect locally on $\{Z_j\}_{j=1}^{m+n}$, to obtain the score function $s(z)$ defined as in~\eqref{eq1}.
%AdaDetect's score function $s$. 
%and uses it for HopSkipJump.

%eliminating the need to train any surrogate classifier.

% \medskip
% We start with describing our attack scheme as follows. 

\textbf{Step 1: Initial detection and BH labeling.} Using the training data $\{Z_j\}_{j=1}^n$ and the mixed sample $\{Z_j\}_{j=k+1}^{n+m}$, form the dataset $\mathcal{D} = \{(Z_i, Y_i)\}_{i=1}^{n+m}$, where $Y_i = 0$ for $i \in [1:k]$ and $Y_i = 1$ for $i \in [k+1:n+m]$ using the positive-unlabeled (PU) framework. Train the score function $s(z) \leftarrow \mathrm{TrainScoreFunction}(\mathcal{D})$. Note that $s(z)$ automatically satisfies the condition in \eqref{eq:score} as $Y_i$ for $i \in [k+1:n+m]$ are the same. Compute empirical $p$-values for $i\in [1:m]$, 
  \begin{equation*}  
    \hat p_i = \frac{1 + \sum_{j=k+1}^n \mathbf{1}\{s(Z_j) \ge s(Z_{n+i})\}}{n-k + 1}.
\end{equation*}
Then apply the BH-procedure to $(\hat p_1, \ldots, \hat p_m)$ to get BH threshold $\hat\tau$ at target level $\alpha$ and produce binary labels
\begin{equation*}
    (\hat Y_1,\dots,\hat Y_m) = \mathrm{BH}\left((\hat p_1, \ldots, \hat p_m), \alpha\right),
\end{equation*}
  where $\hat Y_i = 1$ indicates rejection (detected as non-null) and $\hat Y_i = 0$ indicates non-rejection (undetected).

 \textbf{Step 2: Attack set selection.} Within the set of test samples (i.e., with indices from $[n+1: n+m]$), select a subset $\{Z_{n+i}: i\in \Ac\}$ from the unrejected test samples as the attack target. We set the attack size as (1) {\bf fixed size} where $|\Ac|=m_a$ for some fixed number $m_a$, or (2) {\bf random size} with size $m_{\Ac} = \lfloor \gamma(m-R)\rfloor$, where $\gamma \in (0,1]$ is an ``attack intensity" parameter specified by the attacker. However, we have not implemented this random set setting in our experiments. 
 
% $\Ac$ has a fixed size $m_a$.
 
 %We set the attack size as (1) {\bf fixed size} where $|\Ac|=m_a$ for some fixed number $m_a$, or (2) {\bf random size} with size $m_{\Ac}$.
 
 %, where $\gamma \in (0,1]$ is an ``attack intensity" parameter specified by the attacker. 

%, where the cardinality of the attack set is $|\Ac|=m_{\Ac}$. 

%\begin{equation*}
%\ph_{(\tau_{\text{BH}}+1)}, \,,..., \,\ph_{(\tau_{\text{BH}}+m_{\Ac})}
%\end{equation*}

%\footnote{Our proofs in this work hold for any selection of such a set. See Remark~\ref{rem:select} for one natural and effective way to select the set, as demonstrated in our experiment section.}.

 \textbf{Step 3: Decision-based adversarial perturbation.} For each $i\in \Ac$, generate
  \begin{align}
    \widetilde{Z}_{n+i} &=f_{\text{attack-decision}}(Z_{n+i};\,s(z))\\
    &:=f_{\text{attack-decision}}( Z_{n+i}; \,\{Z_{1},..., Z_n, Z_{n+j}: j\in  \Hc_0\setminus \Ac\},(Z_{n+j}:j\in A\cup \Hc_1))\label{eq:attack2}
  \end{align} 
  such that $\mathbf{1}\{s(Z_{n+i})\ge0.5\}\ne \mathbf{1}\{s(\widetilde{Z}_{n+i})\ge0.5\}$, meaning that the decision is altered.
  % We write 
  % \begin{equation*}
  %     \{Z_{k+1},..., Z_n, Z_{n+j}: j\in \Hc_0\setminus \Ac\}
  % \end{equation*}      as an \emph{unordered} set to highlight that $f_{\text{attack}}$ does not depend on the order of elements in this set.   

 \textbf{Step 4: Applying AdaDetect on the attacked data.} After the attack, the user applies AdaDetect and computes the score function as the first step. As the data is now changed by the attacker, we denote the score function after the attack by $\tilde{s}(z)$, and the empirical $p$-value after the attack by $\pt_i$ for $i\in[1:m]$. %We stress that $\tilde{s}$ still satisfies~\eqref{eq:score}. 

\medskip

The attack set $\Ac$ is inherently random because it depends on the outcome of the BH procedure in Step~1, which in turn depends on the computed $p$-values of the random test samples $\{Z_{n+i}\}_{i=1}^m$. More specifically, each $p$-value $\hat{p}_i$ relies on the entire dataset, including both the training samples $\{Z_j\}_{j=1}^n$ and test samples $\{Z_{n+j}\}_{j=1}^m$, through the score function computation and ranking procedure. In other words, $\Ac$ is a complex yet deterministic function of the complete dataset. 

%Consequently, this randomness propagates through all subsequent analysis: the cardinality $m_{\Ac} = |\Ac|$ is a random variable, and the intersection $\Ac \cap \Hc_0$ with the true null set is random. All of our results that involve $\Ac$ account for this randomness by considering the joint distribution of the whole dataset that induces the distribution of $\Ac$.

% \begin{remark}
% \label{rem:select}

% \end{remark}
\begin{proposition}
\label{prop:perm2}
$f_\text{attack-decision}(\cdot\,;\, g(z))$ does not depend on the order of elements in $\{Z_{k+1},..., Z_n, Z_{n+j}: j\in \Hc_0\setminus \Ac\}$.
\end{proposition}

This proposition captures the main subtlety difference between this attack scheme and the oracle setting in Theorem~\ref{thm:1}. It holds because $f_\text{attack-decision}(\cdot\,;\, g(z))$ only relies on the score function $s(z)$, and $s(z)$ is invariant to order of elements in $\{Z_{k+1},\dots,Z_{n+m}\}$ according to~\eqref{eq:score}, as a consequence of the PU framework.

%\subsection{Analysis}
% We denote the corresponding $\text{FDR}$ as  $\text{FDR}^*_{\text{attack--decision}}$, and quantify the loss in FDR caused by the attack.
\begin{proposition}
\label{prop:2}
    Consider that $\Ac$ is a fixed set with $m_{a}=|\Ac|$. Under Assumption~\ref{ass1}, with the score function $\tilde{s}$ satisfying the permutation invariance property in~\eqref{eq:score} and the attack scheme $f_\text{attack-decision}(\cdot\,;\, g(z))$ being order-invariant as in~\eqref{eq:attack2}, the FDR after the attack, denoted by $\text{FDR}^*_{\text{attack--decision}}$ can be upper bounded as
    \begin{equation}
    \label{eq:prop3}
\text{FDR}^*_{\text{attack--decision}} \le \frac{m_0}{m}\a +m_{a}\cdot\mathbb{E}\left[\frac{1}{\Rt_{m_a}\lor 1} \right],
    \end{equation}where the expectations are taken over the randomness in the training and test samples $\{Z_j\}_{j=1}^{m+n}$. 

When $\Ac$ is random with fixed size $|\Ac|=m_a$, we have $\text{FDR}_{\text{attack}} \le \text{FDR}^*_{\text{attack--decision}}$. 

Furthermore, we have $\text{FDR}_{\text{attack}} \le \text{FDR}^*_{\text{attack--decision}}\le \alpha$ as long as
\begin{equation}
    h(m_a)\le \frac{m_1}{m}\,\alpha\,\quad\text{ where }h(x):=m_a\,\cdot\mathbb{E}\left[\frac{1}{\Rt(m_a)\lor 1}\right].
\end{equation}

\end{proposition}

% m_a\mathbb{E}\left[\frac{1}{\Rt(m_a)\lor 1} \right]\le \alpha\frac{m_1}{m}. 
% \] 
    
\begin{remark}
    The reason we have $m_0/m$ in~\eqref{eq:prop3} rather than $(m_0-m_a)/m$ as in Theorem~\ref{thm:1} is because $|\Hc_0\setminus\Ac|$ may be strictly bigger than $m_0-m_a$ but still bounded by $m_0$, while it is exactly equal to $m_0-m_a$ in the oracle setting.

\end{remark}

This result follows directly from proof of Theorem~\ref{thm:1}, as the only difference between Theorem~\ref{prop:2} and Proposition~\ref{thm:1} is that the score function is trained differently. But according to Proposition~\ref{prop:perm2}, the score function for attack is still invariant under the permutation of $\{Z_{k+1},..., Z_n, Z_{n+j}: j\in  \Hc_0\setminus \Ac\}$, which implies that the ``matching" of the invariance between $f_\text{attack-decision}(\cdot\,;\, g(z))$ and $\tilde{s}(z)$ still holds. Another way to see this is that because both the attacker and user apply PU, and thus assign the same labels to $\{Z_{k+1},..., Z_n, Z_{n+j}: j\in  \Hc_0\setminus \Ac\}$. This makes the rest of the proof exactly the same as that for Theorem~\ref{thm:1}. 

When the size of the attack set is random, denoted by $m_{\Ac}$, it will show up inside the expectation as follows.
\begin{corollary}
    Consider that $\Ac$ is random with a random size $|\Ac|=m_{\Ac}$. Under Assumption~\ref{ass1}, with~\eqref{eq:score} and~\eqref{eq:attack}, the FDR after the attack is
    \begin{equation}
\text{FDR}_{\text{attack}} \le \frac{m_0}{m}\a +\mathbb{E}\left[\frac{m_{\Ac}}{\Rt(m_{\Ac})\lor 1} \right],
    \end{equation}
    where the expectations are taken over the randomness in the training and test samples $\{Z_j\}_{j=1}^{m+n}$, which induces randomness in $\Ac$ and $\Rt(m_{\Ac})$.
\end{corollary}

\subsection{Attack set selection}
\label{app:power}

It has been proved that AdaDetect has a strong detection power (the probability of correctly rejecting a non-null), as shown in~\cite[Theorem~5.1]{marandon2024adaptive}. This implies that, in the surrogate method, the set $\Ac$ will nearly contain all indices from true nulls because the non-nulls are mostly rejected. We use this to show that $\P(\Ac\,\subseteq\, \Hc_0)$ is high. Instead of showing detailed technical steps following the proof~\cite[Theorem~5.1]{marandon2024adaptive}, we choose to provide an informal argument to connect the power of AdaDetect and $\P(\Ac\,\subseteq\, \Hc_0)$.

% the upper bound $\mathbb{E}[\frac{m_{\Ac}}{\Rt_{m_a}\lor 1}]$ in~\eqref{thm:1} is relatively tight (see details in Proposition~\ref{pro:upper}).

% In the following proposition, we show that the upper bound in~\eqref{eq:up} is relatively tight by making a connection between step (a) in~\eqref{eq:up} and the power of AdaDetect. Roughly speaking, the upper bound is relatively tight when the power of AdaDetect is decent. 

\begin{proposition}
\label{pro:upper}
Under the assumptions in~\cite[Theorem~5.1]{marandon2024adaptive} and when $\Ac$ is randomly selected from the unrejected indices with $m_{\Ac}$, we have that for some small $\delta'$ and $\eta'$, 
\begin{equation}
     P\left(\Ac\,\subseteq\, \Hc_0\right)\ge (1-\delta')\cdot l(\eta'),
\end{equation}
   where $l(\eta')\to 1$ as $\eta'\to 0$. 
\end{proposition}

According to  from~\cite[Theorem~5.1]{marandon2024adaptive}, we have that the rejection set by AdaDetect at level $\l$, denoted by $\text{AdaDetect}_{\l}$ satisfies  
\begin{align}
\P\biggl(\frac{|\text{AdaDetect}_{\l}\cap \Hc_1|}{m_1}\ge 1-\eta\biggr)\ge 1-\delta, \label{eq:5.1}
\end{align}
for some small $\delta$ and $\eta$. This can be adapted to our setting as follows, with one approximation, where we treat the data samples being attacked as non-nulls. After the attack, we have
\begin{align}
\P\biggl(\frac{|\widetilde{\text{AdaDetect}}_{\l}\cap (\Hc_1\cup \Ac)|}{|\Hc_1\cup \Ac|}\ge 1-\eta'\biggr)\ge 1-\delta', 
\end{align}
where $\widetilde{\text{AdaDetect}}_{\l}$ denotes the rejection set after the attack, and we note that the number of non-nulls becomes $|\Hc_1\cup \Ac|\le m_1+m_{\Ac}$ after the attack. 

Since $\widetilde{\Rc}=\widetilde{\text{AdaDetect}}_{\l}$ and $|\widetilde{\Rc}\cap (\Hc_1\cup\Ac)| = \Rt_{m_a}-\Vt$, we have that the unrejected set contains $(|\Hc_1\cup \Ac|- (\Rt_{m_a}-\Vt))) \le |\Hc_1\cup \Ac|\cdot \eta'$ with probability at least $1-\delta'$. Define $\Ec \;=\; \left\{\,\big|\widetilde R - \widetilde V\big| \;\ge\; |\Hc_1\cup \Ac|\cdot(1-\eta')\right\}$. We use the shorthand $Z:=\{Z_i\}_{i=1}^{n+m}$ to denote the whole dataset. Given any fixed dataset $Z=z$, the only remaining randomness comes from the random selection (and it is independent of $Z$), while the random variables $\widetilde R$, $\Ac$, and $m_{\Ac}$ take on realizations as    $\widetilde R(z)$, $\Ac(z)$, and $m_{\Ac(z)}$, respectively. We now have 
\begin{align*}
&P\left(\Ac\subseteq \Hc_0\right)\ge P(\Ec)\cdot\;P\left(\Ac\subseteq \Hc_0\mid \Ec\right)\\
&\hspace{0.5em}\ge (1-\delta')\;\cdot\int_{z}
\frac{\displaystyle\binom{\,m-\widetilde R(z) - |\Hc_1\cup \Ac(z)|\cdot\eta'\,}{m_{\Ac(z)}}}
     {\displaystyle\binom{\,m-\widetilde R(z)\,}{m_{\Ac(z)}}}
     \cdot P(Z=z|\Ec)\, dz\,
     :=(1-\delta')\cdot l(\eta'),
\end{align*}
where $l(\eta')\to 1$ as $\eta'\to 0$. 
The last step follows since given event $\Ec$ and $\{Z=z\}$, we have $|\widetilde{\Rc}(z)\cap (\Hc_1\cup\Ac(z))|\ge |\Hc_1\cup \Ac(z)|\cdot(1-\eta')$, which implies the number of unrejected non-nulls is smaller than $|\Hc_1\cup \Ac(z)|\cdot\eta'$.

\section{Comparison between our surrogate algorithm and INCREASE-c~\cite{chen2024adversarial}}\label{app:compare}
The two approaches are fundamentally different. We focus on attacking data samples directly, while INCREASE-c perturbs $p$-values. From a practical perspective, the crucial difference between the two lies in the perturbation strategy. While our method targets data point near the learned decision boundary via a surrogate score function, INCREASE-c effectively seeks to alter $p$-values that are often far from the decision boundary to ensure a collective shift in the empirical FDR. Consequently, INCREASE-c requires a significantly larger per-sample perturbation to force a change in the rejection threshold, whereas our approach exploits the local sensitivity of the score function to induce misclassification with minimal data distortion.
We focus on credit card dataset as in \cite{chen2024adversarial} and the BH threshold in our surrogate approach is about $0.0083$, and we are perturbing the data with $p$-values that are slightly above this threshold. 

To evaluate the comparative impact, we replicate the INCREASE-c experimental framework on the same credit card dataset. An isolation forest is trained with a training sample selected uniformly at random from the set of true nulls; a calibration subset of strictly genuine transactions is randomly selected from the null data, after which a test sample is formed by combining the remaining null data with a random subset of fraudulent transactions. The test sample is transformed to $p$-values, and the BH procedure (with a level of $0.1$) is applied to identify the set of fraudulent transactions. (The full details of the experiment can be found in~\cite{chen2024adversarial}.) We execute INCREASE-$c$ against these $p$-values and their ground-truth labels across $100$ simulations for each attack budget $c = 1, 5, 10, 20$. The results are reported in Table~$6$. The high average value of the selected $p$-values (i.e., $\mathbb{E}[p_{\text{selected}}]$) confirms that INCREASE-c targets data points far from the (average) BH cutoff (i.e., the last column in Table~IX). Due to the drastic change in perturbed $p$-values, there is a clear impact on the resulting FDR and total rejection count. 
\begin{table}[ht]
\centering
\caption{Experiment E.1: INCREASE-c results: perturbation and cutoff}
\begin{tabular}{ccccccc}
\toprule
$c$ & \begin{tabular}{@{}c@{}}original \\ FDR\end{tabular} & \begin{tabular}{@{}c@{}}INCREASE-c \\ FDR\end{tabular} & \begin{tabular}{@{}c@{}}original \\ $\mathbb{E}[R]$\end{tabular} & \begin{tabular}{@{}c@{}}INCREASE-c \\ $\mathbb{E}[R]$\end{tabular} & $\mathbb{E}[p_{\text{selected}}]$ & \begin{tabular}{@{}c@{}}average \\ BH cutoff\end{tabular}\\ \midrule
1  & 0.08 & 0.13 & 43.52 & 45.79 & 0.99 & 0.0052\\
5  & 0.09 & 0.17 & 55.24 & 64.39 & 0.99 & 0.0061\\
10 & 0.08 & 0.24 & 46.33 & 64.00 & 0.99 &0.0045\\
20 & 0.08 & 0.31 & 55.48 & 90.28 & 0.98 &0.0034\\ \bottomrule
\end{tabular}
\end{table}

\section{Additional Experiments}
\label{app:exp}

\noindent{\bf Synthetic data generation:}
We generate two types of data: null samples from distribution $P_0$ and non-null samples from distribution $P_1$. We let $d=20$ for all the synthetic data.

\textbf{Independent Gaussian:} We consider $P_0 = \mathcal{N}(0, I_d)$ and $P_1 = \mathcal{N}(\mu, I_d)$, where $\mu \in \mathbb{R}^d$ is a sparse mean shift vector: the first five coordinates are set to $\sqrt{2\log(d)}$ and the remaining coordinates are zero. 

 \textbf{Non-Gaussian:}  We let the first two coordinates of nulls and non-nulls be drawn independently from Beta distributions: $P_0:  (X_1, X_2) \sim \text{Beta}(5,5)$ and $P_1:  (X_1, X_2) \sim \text{Beta}(1,3)$. The remaining coordinates are drawn i.i.d. from $\text{Beta}(1,1)$ under both $P_0$ and $P_1$.

 \textbf{Exchangeable Gaussian:} Let $T = \mathcal{N}(\mu, \Sigma)$ be the $d$-variate Gaussian distribution with mean vector 
$\mu = [\mu_1,\dots,\mu_d]^\top$ and covariance matrix $\Sigma = [\sigma_{ij}]_{i,j=1}^d$. 
Suppose $T$ is exchangeable, i.e.,
\begin{equation*}
\mu_i = \mu_j =: a, 
\sigma_{ii} = \sigma_{jj} =: b^2, 
\sigma_{ij} = \sigma_{kl} =: c,
\end{equation*}
for all $i,j,k,l$ with $i\neq j$ and $k\neq l$. 
Then the covariance matrix can be written as
\begin{equation*}
\Sigma = c \mathbf{1}\mathbf{1}^\top + (b^2 - c) I_d,\quad \text{ where }\mathbf{1} = [1,\dots,1]^\top \in \mathbb{R}^d.
\end{equation*}

We define the null and non-null distributions as $P_0 = \mathcal{N}(a\mathbf{1}, \Sigma)$ and $P_1 = \mathcal{N}((a+\delta)\mathbf{1}, \Sigma)$, where $\delta > 0$ introduces a mean shift across all coordinates. 
Thus $P_0$ and $P_1$ share the same exchangeable covariance structure but differ in their mean vectors.

\smallskip
\noindent{\bf Experiment~A.1: Mismatched score function configurations.}
We investigate both oracle and surrogate attack performance when using different model architectures and parameters for the score functions, with attack size $m_a=200$. This experiment comprises distinct configurations for each attack type:
\begin{itemize}
    \item \textbf{RF--NN:} AdaDetect score function $\tilde{s}(z)$ uses RF, attacker score function ${g}(z)$ uses NN.
    \item \textbf{RF--RF:} Both score functions use RFs with the same hyperparameters.
\end{itemize}

{\renewcommand{\arraystretch}{1.5}
\begin{table*}[h!]
\centering
\caption{Experiment~A.2: FDR + RF-NN}
\renewcommand{\arraystretch}{1.0}
\begin{tabular}{lcccc}
\hline
\hline
{\bf Dataset} & {\bf Independent Gaussian} & {\bf Non-Gaussian} & {\bf Exchangeable Gaussian} \\
\hline\hline
original FDR    &$0.08 \pm 0.03$  & $0.08 \pm 0.04$ &$0.08 \pm 0.04$   \\
  \hline
oracle ($m_a=200$)          & $0.66 \pm 0.00$ &$0.70 \pm 0.02$   & $0.67 \pm 0.00$    \\
surrogate ($m_a=200$)& $0.68 \pm 0.00$ &$0.63 \pm 0.01$   & $0.65 \pm 0.02$   \\
\hline
estimated upper bound& $0.77$ &$0.76$   & $0.67$ \\
\hline\hline
\end{tabular}

\end{table*}
}
{\renewcommand{\arraystretch}{1.5}
\begin{table*}[h!]
\centering
\caption{Experiment~A.3: FDR + NN}
\renewcommand{\arraystretch}{1.0}
\begin{tabular}{lcccc}
\hline
\hline
{\bf Dataset} & {\bf Credit-card} & {\bf Shuttle} & {\bf KDD} & {\bf Mammography} \\
\hline\hline
original FDR   &$0.09 \pm 0.05$  &$0.01 \pm 0.01$  & $0.02 \pm 0.01$ &$0.09 \pm 0.05$  \\
  \hline
oracle+{\bf hop.}          &$0.67 \pm 0.04$  &$0.44 \pm 0.01$  &  $0.59 \pm 0.03$ &$0.78 \pm 0.01$ \\
surrogate+{\bf hop.} &$0.67 \pm 0.05$  &$0.43 \pm 0.00$  & $0.61 \pm 0.02$  &$0.65 \pm 0.02$ \\
oracle+{\bf bound.}          &$0.66 \pm 0.02$  & $0.36 \pm 0.05$ &  $0.47 \pm 0.06$&$0.64 \pm 0.05$ \\
surrogate+{\bf bound.} &$0.65 \pm 0.02$  &$0.45 \pm 0.09$  &$0.43 \pm 0.05$  &$0.61 \pm 0.04$ \\
\hline
estimated upper bound& $0.77$ &$0.76$   & $0.81$ & $0.80$   \\
\hline\hline
\end{tabular}

\end{table*}
}

{\renewcommand{\arraystretch}{1.5}
\begin{table*}[h!]
\centering
\caption{Experiment~A.3: Power + NN}
\renewcommand{\arraystretch}{1.0}
\begin{tabular}{lcccc}
\hline
\hline
{\bf Dataset} & {\bf Credit-card} & {\bf Shuttle} & {\bf KDD} & {\bf Mammography} \\
\hline\hline
original power    & $0.80 \pm 0.03$   & $0.84 \pm 0.09$  &  $0.78 \pm 0.04$ &$0.53 \pm 0.09$  \\
  \hline
  oracle+{\bf hop.} & $0.95 \pm 0.03$ &$0.98 \pm 0.01$  & $0.88 \pm 0.02$ & $0.65 \pm 0.01$ \\
surrogate+{\bf hop.}          &$0.86 \pm 0.04$  & $0.99 \pm 0.01$ & $0.86 \pm 0.03$ &$0.87 \pm 0.01$  \\

oracle+{\bf bound.}          &$0.93 \pm 0.03$  & $0.94 \pm 0.01$ & $0.99 \pm 0.01$ &$0.77 \pm 0.06$ \\
surrogate+{\bf bound.} & $0.95 \pm 0.02$ & $0.99 \pm 0.01$ &$0.97 \pm 0.01$  & $0.80 \pm 0.07$\\
\hline\hline
\end{tabular}
\end{table*}
}

\smallskip
\noindent{\bf Experiment A.2: Real-world data with NN models.} 
This experiment employs NN architectures for both score functions on the four real-world datasets, evaluating both oracle and surrogate attack performance with attack size $m_a=200$. This allows us to evaluate how each attack type performs when both the target model and attacker use NN-based approaches on realistic data distributions. We compare boundary attack with our default HSJA attack under both oracle and surrogate attack schemes, using identical RF architectures for both score functions with $m_a=200$. The result shows that both HSJA and boundary attack are successful at increasing the FDR for NN models in real-world data.

\end{document}